\documentclass[11pt]{article}

\usepackage[preprint]{acl}
 \usepackage{amsmath}

\usepackage{times}
\usepackage{latexsym}
\usepackage{microtype}
\usepackage{booktabs}
\usepackage{tabularx}
\usepackage{array}
\usepackage{colortbl} 
\newcolumntype{Y}{>{\centering\arraybackslash}X}
\newcolumntype{L}{>{\raggedright\arraybackslash}X}
\newcolumntype{C}{>{\centering\arraybackslash}X}

\usepackage{tcolorbox}
\tcbuselibrary{skins}
\tcbuselibrary{breakable}
\tcbuselibrary{raster}
\tcbuselibrary{hooks}
\tcbuselibrary{theorems}
\tcbuselibrary{listings}
\tcbuselibrary{documentation}
\usepackage{enumitem}
\definecolor{blue}{HTML}{5989cf}
\definecolor{sub}{HTML}{cde4ff}
\newtcolorbox{todobox}{
    colback = sub, 
    colframe = blue, 
    boxrule = 0pt, 
    leftrule = 6pt 
}
\tcbset{
  findingbox/.style={
    colback=gray!5,             
    colframe=black!30,          
    coltitle=white,             
    colbacktitle=black!60,      
    fonttitle=\small\bfseries,      
    boxrule=0.3pt,              
    arc=2pt,                    
    left=6pt, right=6pt, top=2pt, bottom=2pt, 
    enhanced,
    boxed title style={size=small, colframe=black!60, colback=black!60, sharp corners,
    top=1pt, bottom=1pt, left=2pt, right=2pt
    },
  }
}

\usepackage[T1]{fontenc}

\usepackage[utf8]{inputenc}

\usepackage{microtype}

\usepackage{inconsolata}

\usepackage{graphicx}

\title{The Anatomy of an Edit: Mechanism-Guided Activation Steering for Knowledge Editing}

\author{
Yuan Cao$^{1}$ \hspace*{0.2cm}
{Mingyang Wang$^{2,3}$} \hspace*{0.2cm} {Hinrich Sch\"{u}tze$^{2,3}$} \\
  $^1$Technical University of Munich \hspace*{0.2cm}
  $^2$LMU Munich \hspace*{0.2cm} \\
  $^3$Munich Center for Machine Learning (MCML) \\
  \texttt{trillionyuan.cao@tum.de}
  }

\begin{document}
\maketitle

\begin{figure*}[t]
    \centering
    \includegraphics[
        height=5.5cm,
        keepaspectratio,
        trim=15 10 15 10,
        clip
    ]{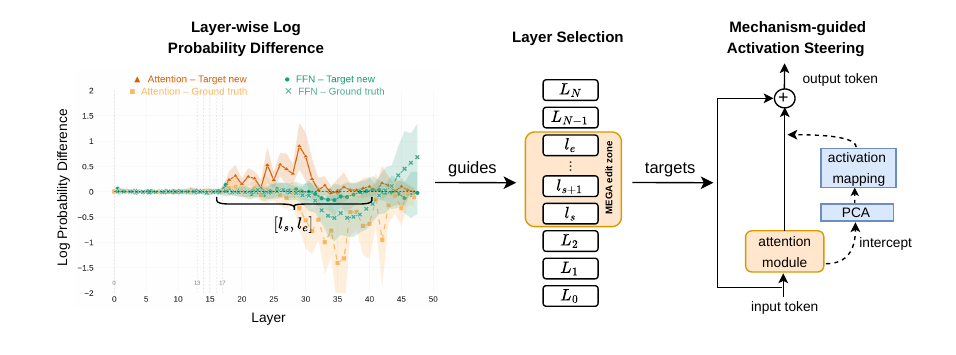}
    \caption{
    \textbf{Overview of MEGA.} \textbf{(Left):} Post-edit NLKA identifies mid-to-late attention layers with high leverage on the target vs. original token. \textbf{(Middle):} MEGA uses this signal to select an edit zone and \textbf{(Right):} applies PCA-stabilized attention–residual steering at inference time, promoting the edited fact without changing weights.
    }
    \label{fig:teaser}
\end{figure*}

\begin{abstract}
Large language models (LLMs) are increasingly used as knowledge bases, but keeping them up to date requires targeted knowledge editing (KE). However, it remains unclear how edits are implemented inside the model once applied. In this work, we take a mechanistic view of KE using neuron-level knowledge attribution (NLKA). Unlike prior work that focuses on pre-edit causal tracing and localization, we use post-edit attribution -- contrasting successful and failed edits -- to isolate the computations that shift when an edit succeeds. Across representative KE methods, we find a consistent pattern: mid-to-late attention predominantly promotes the new target, while attention and FFN modules cooperate to suppress the original fact.
Motivated by these findings, we propose \textbf{MEGA}, a \textbf{ME}chanism-\textbf{G}uided \textbf{A}ctivation steering method that performs attention-residual interventions in attribution-aligned regions without modifying model weights. On CounterFact and Popular, MEGA achieves strong editing performance across KE metrics on GPT2-XL and LLaMA2-7B.
Overall, our results elevate post-edit attribution from analysis to engineering signal: by pinpointing where and how edits take hold, it powers MEGA to deliver reliable, architecture-agnostic knowledge edits.
\end{abstract}

\section{Introduction}
\label{sec:intro}

Large language models (LLMs) are increasingly used as knowledge bases, but their internal facts can be outdated or application-specific. Knowledge editing aims to update specific factual associations without retraining from scratch or degrading unrelated behavior. Existing approaches include weight-editing methods such as ROME and MEMIT \citep{meng2023locatingeditingfactualassociations,meng2023masseditingmemorytransformer}, fine-tuning and parameter-efficient adaptation \citep{hu2021loralowrankadaptationlarge}, in-context editing \citep{zheng2023editfactualknowledgeincontext}, and activation steering methods \citep{scialanga2025sakesteeringactivationsknowledge}. 

Despite progress, many methods choose edit layers based on pre-edit localization with causal tracing, i.e., where a fact appears to originate in the base model. However, such localization has been shown to be a poor predictor of effective intervention sites \citep{hase2023does}. Instead, we argue that analyzing post-edit attribution can reveal which components actually implement the updated knowledge and directly inform intervention design.

We take a mechanistic view of knowledge editing on GPT2-XL and LLaMA2-7B. Using neuron-level knowledge attribution (NLKA) \citep{yu2024neuronlevelknowledgeattributionlarge}, we compare four representative methods—ROME, MEMIT, FT, and in-context editing—by tracking how layer- and component-wise contributions to target-token log-probabilities shift before and after editing. Across methods, we observe a consistent signature: (i) attention and FFNs jointly suppress the original fact; (ii) mid-to-late attention predominantly promotes the new target; and (iii) late residual streams act as high-leverage aggregation points, as suggested by attribution trends.

Guided by these findings, we propose \textbf{MEGA}, a mechanism-guided activation steering method that performs attention-residual interventions in attribution-aligned regions. As shown in Figure~\ref{fig:teaser}, rather than relying on heuristic last-layer edits, MEGA targets the mid-to-late layers where attribution and ablations indicate maximal leverage. On CounterFact and Popular, MEGA yields strong edits across GPT2-XL and LLaMA2-7B, while FFN-only steering shows limited gains.

Overall, our contributions are as follows. 
First, \textbf{post-edit knowledge attribution}: we present a unified post-edit NLKA analysis across four KE methods and two LLM families, revealing consistent component roles and contrasting with prior pre-edit attribution approaches. 
Second, \textbf{mechanism-guided activation steering}: we introduce MEGA, an attention-residual steering method aligned with attribution-identified control regions, achieving strong edits without modifying model weights. 
Finally, \textbf{bridging analysis and intervention}: we show that post-edit attribution is not merely diagnostic but actionable, enabling mechanism-aware editing strategies that generalize across LLMs.

\begin{figure*}[t]
\centering
\begin{tabular}{cc}
\includegraphics[width=8cm]{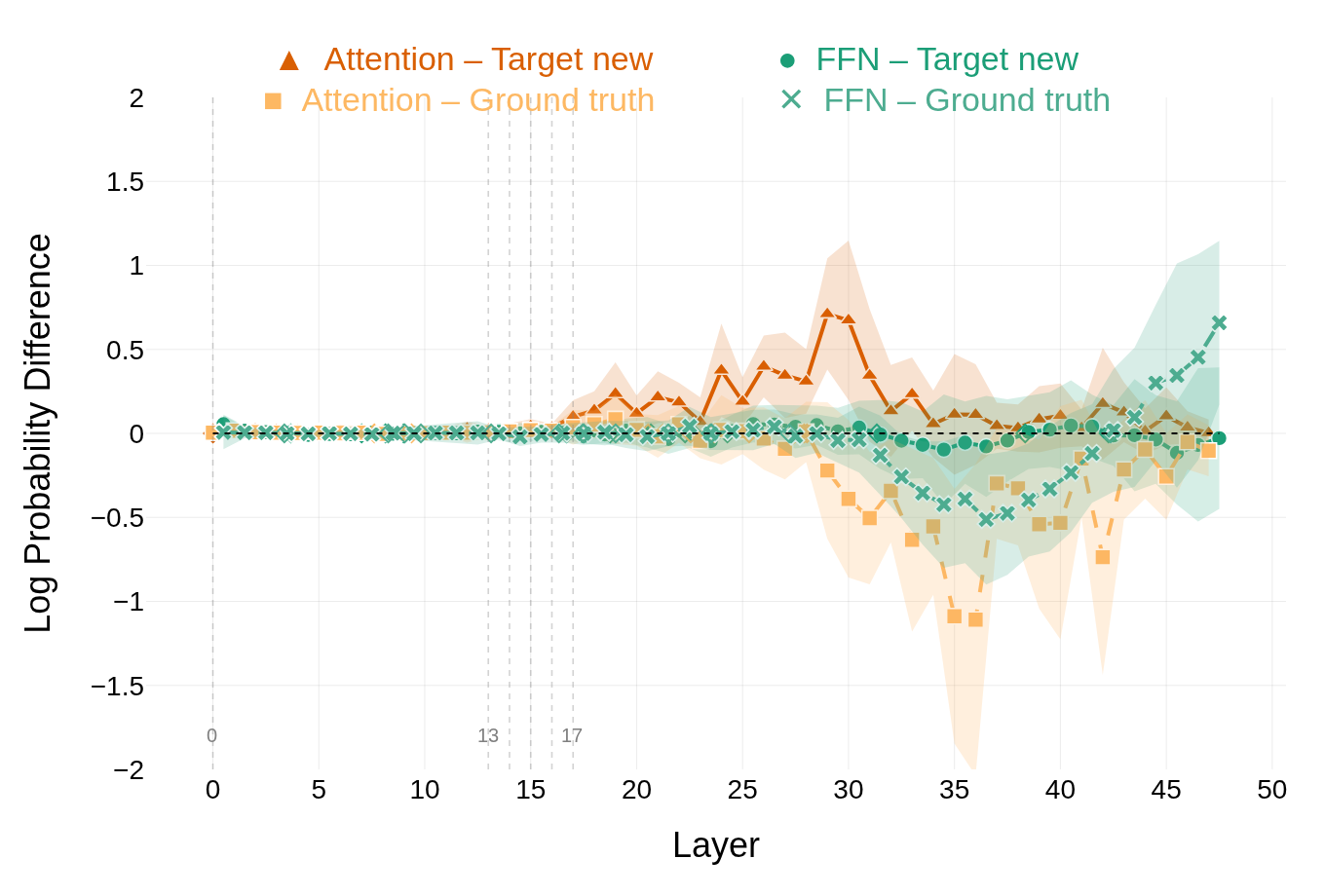} &
\includegraphics[width=8cm]{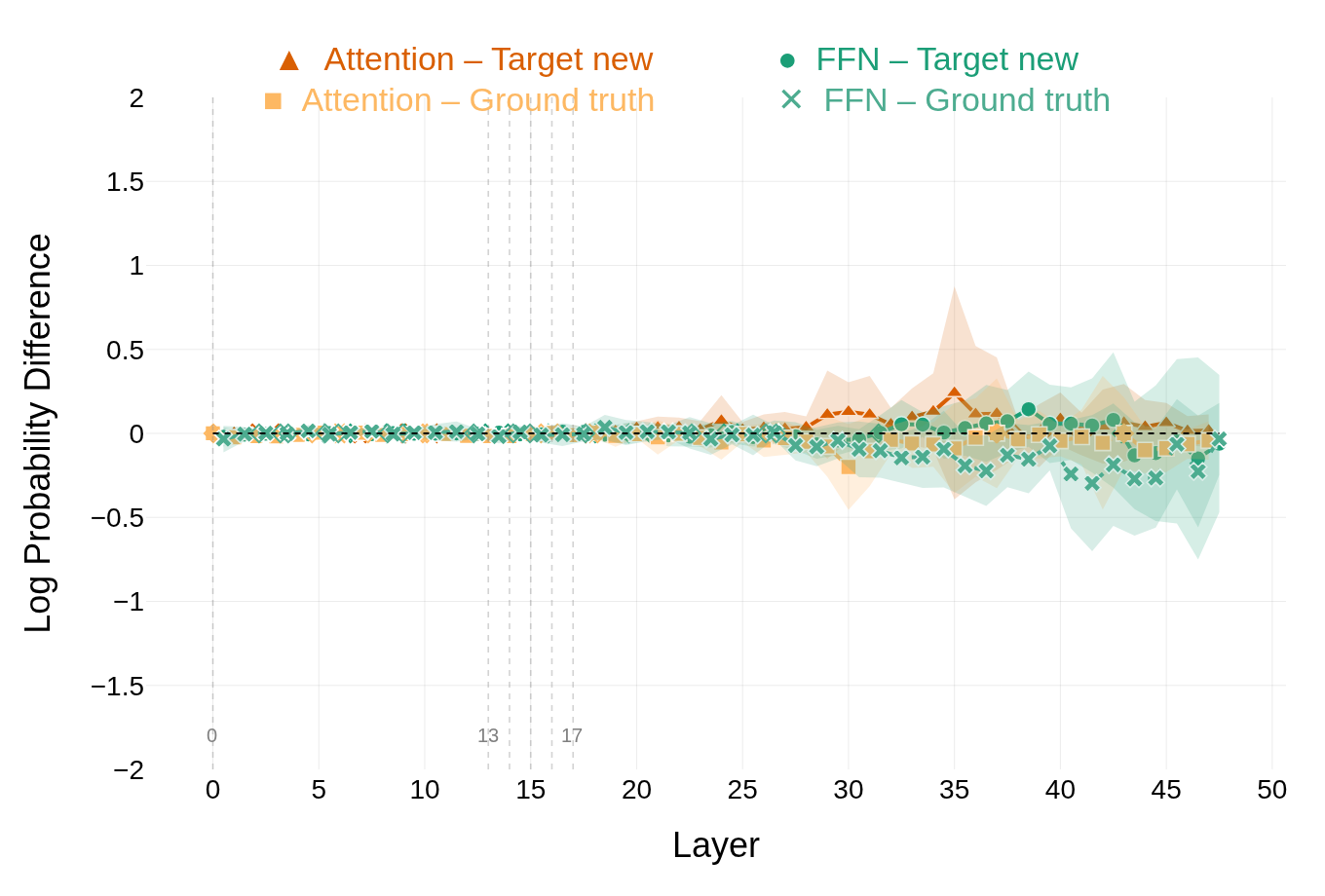} \\
(a) Success (L13--17) & (b) Failure (L13--17)
\end{tabular}
\caption{\textbf{MEMIT success vs.\ failure.} Mean contribution differences across multiple cases; positive promotes new target, negative suppresses original.}
\label{fig:cf-gpt2-memit}
\end{figure*}

\begin{table*}[h]
\centering
\caption{Best-performing methods on CounterFact for GPT2-XL (left) and LLaMA2-7B (right). We report accuracy (Acc), generalization (Gen), specificity (Spec), and two doubt-robustness metrics (DI, DII), plus their average (Avg).}
\label{tab:cf_best_of}
\footnotesize
\setlength{\tabcolsep}{3pt}
\renewcommand{\arraystretch}{1.0}

\begin{minipage}{0.48\textwidth}
\centering
\textbf{GPT2-XL}
\begin{tabularx}{\columnwidth}{@{}>{\raggedright\arraybackslash}p{1.5cm} c c Y Y Y Y Y Y@{}}
\toprule
\textbf{Config} & \textbf{Lyrs} & \textbf{PCA}
& \textbf{Acc} & \textbf{Gen} & \textbf{Spec} & \textbf{DI} & \textbf{DII} & \textbf{Avg}\\
\midrule

FT    & 0      & --  & 0.97 & 0.09 & 0.80 & 0.17 & 0.03 & 0.41 \\
ROME  & 17     & --  & 1.00 & 0.80 & 0.86 & 0.36 & 0.07 & 0.62 \\
MEMIT & 13--17 & --  & 0.81 & 0.45 & 0.99 & 0.25 & 0.03 & 0.51 \\
IKE   & --     & --  & 0.99 & 0.70 & 0.17 & 0.06 & 0.01 & 0.39 \\
SAKE  & 47     & --  & 0.99 & 0.85 & 0.83 & 0.99 & 0.99 & 0.93 \\
\midrule
\rowcolor{gray!12}
\textbf{MEGA} & 15--40 & 512
& 0.99 & 0.93 & 0.87 & 1.00 & 1.00 & \textbf{0.96} \\
\bottomrule
\end{tabularx}
\end{minipage}
\hfill
\begin{minipage}{0.48\textwidth}
\centering
\textbf{LLaMA2-7B}
\begin{tabularx}{\columnwidth}{@{}>{\raggedright\arraybackslash}p{1.5cm} c c Y Y Y Y Y Y@{}}
\toprule
\textbf{Config} & \textbf{Lyrs} & \textbf{PCA}
& \textbf{Acc} & \textbf{Gen} & \textbf{Spec} & \textbf{DI} & \textbf{DII} & \textbf{Avg}\\
\midrule

FT    & 21     & --  & 0.45 & 0.43 & 0.44 & 0.01 & 0.66 & 0.40 \\
ROME  & 5      & --  & 1.00 & 0.81 & 0.91 & 0.03 & 0.41 & 0.63 \\
MEMIT & 4--8   & --  & 1.00 & 0.83 & 0.94 & 0.12 & 0.54 & 0.69 \\
IKE   & --     & --  & 1.00 & 0.89 & 0.34 & 0.00 & 0.05 & 0.46 \\
SAKE  & 31     & --  & 0.98 & 0.83 & 0.84 & 0.92 & 0.96 & 0.91 \\
\midrule
\rowcolor{gray!12}
\textbf{MEGA} & 13--31 & 512
& 0.99 & 0.91 & 0.89 & 1.00 & 0.97 & \textbf{0.95} \\
\bottomrule
\end{tabularx}
\end{minipage}
\end{table*}

\begin{table*}[t]
\centering
\caption{Best-performing methods on Popular for GPT2-XL (left) and LLaMA2-7B (right). We report accuracy (Acc), compositionality (CI, CII), relation specificity (RS), subject aliasing (SA), and their average (Avg).}
\label{tab:pop_best_of}
\footnotesize
\setlength{\tabcolsep}{3pt}
\renewcommand{\arraystretch}{1.0}

\begin{minipage}{0.48\textwidth}
\centering
\textbf{GPT2-XL}
\begin{tabularx}{\columnwidth}{@{}>{\raggedright\arraybackslash}p{1.5cm} c c Y Y Y Y Y Y@{}}
\toprule
\textbf{Config} & \textbf{Lyrs} & \textbf{PCA}
& \textbf{Acc} & \textbf{CI} & \textbf{CII} & \textbf{RS} & \textbf{SA} & \textbf{Avg}\\
\midrule

FT    & 0       & --  & 0.30 & 0.15 & 0.00 & 0.52 & 0.25 & 0.24 \\
ROME  & 17      & --  & 0.96 & 0.22 & 0.00 & 0.27 & 0.53 & 0.40 \\
MEMIT & 13--17  & --  & 0.55 & 0.24 & 0.00 & 0.38 & 0.22 & 0.28 \\
IKE   & --      & --  & 0.96 & 0.09 & 0.00 & 0.39 & 0.01 & 0.29 \\
SAKE  & 47      & --  & 0.45 & 0.60 & 0.35 & 0.38 & 0.70 & \textbf{0.50} \\
\midrule
\rowcolor{gray!12}
\textbf{MEGA}  & 31--47     & 512 & 0.80 & 0.28 & 0.45 & 0.32 & 0.64 & \textbf{0.50} \\

\bottomrule
\end{tabularx}
\end{minipage}
\hfill
\begin{minipage}{0.48\textwidth}
\centering
\textbf{LLaMA2-7B}
\begin{tabularx}{\columnwidth}{@{}>{\raggedright\arraybackslash}p{1.5cm} c c Y Y Y Y Y Y@{}}
\toprule
\textbf{Config} & \textbf{Lyrs} & \textbf{PCA}
& \textbf{Acc} & \textbf{CI} & \textbf{CII} & \textbf{RS} & \textbf{SA} & \textbf{Avg}\\
\midrule
FT    & 21     & --  & 0.48 & 0.05 & 0.02 & 0.53 & 0.24 & 0.26 \\
ROME  & 5      & --  & 0.99 & 0.20 & 0.04 & 0.53 & 0.73 & \textbf{0.50} \\
MEMIT & 4--8   & --  & 0.98 & 0.18 & 0.03 & 0.41 & 0.79 & 0.48 \\
IKE   & --     & --  & 0.93 & 0.06 & 0.00 & 0.86 & 0.05 & 0.38 \\
SAKE  & 31     & --  & 0.47 & 0.13 & 0.24 & 0.48 & 0.63 & 0.39 \\ 
\midrule
\rowcolor{gray!12}
\textbf{MEGA}       & 13--31     & 512 & 0.82 & 0.07 & 0.25 & 0.33 & 0.64 & 0.42 \\
\bottomrule
\end{tabularx}
\end{minipage}
\end{table*}

\section{Background and Related Work}
\label{sec:background}

\paragraph{Knowledge attribution.}
Knowledge attribution methods aim to localize which internal units or components are responsible for a model’s predictions. NLKA \citep{yu2024neuronlevelknowledgeattributionlarge} provides an observational, post-hoc decomposition of a token’s log-probability into additive contributions across layers and components (e.g., attention and FFN). NLKA measures how each component’s output, when integrated into the residual stream and projected through the final layer norm and LM head, changes the log-probability of a target token. The model itself is not modified, enabling direct comparison between edited and unedited networks and identification of promising intervention targets.

\paragraph{Knowledge editing.}
Knowledge editing aims to update factual knowledge in pretrained language 
models while preserving general capabilities \citep{meng2022locating,
meng2023masseditingmemorytransformer,zhang2024comprehensivestudyknowledgeediting}. Given a fact 
$(x_{\text{src}}, y_{\text{tgt}})$, the objective is to increase the 
likelihood of $y_{\text{tgt}}$ in relevant contexts without affecting 
unrelated predictions.

Prior work spans weight-level editors such as ROME and MEMIT \citep{meng2023locatingeditingfactualassociations,meng2023masseditingmemorytransformer,wang-etal-2025-bring}, FT and parameter-efficient adaptation methods \citep{hu2021loralowrankadaptationlarge,houlsby2019parameterefficienttransferlearningnlp,wang-etal-2024-better}, memory-based and in-context approaches including SERAC, GRACE, and IKE \citep{mitchell2022memorybasedmodeleditingscale,hartvigsen2023aginggracelifelongmodel,zheng2023editfactualknowledgeincontext}, and inference-time activation steering methods such as SAKE \citep{scialanga2025sakesteeringactivationsknowledge}. SAKE learns an optimal-transport map on final-layer residual states and uses a scope detector to apply edits only in scope. Our work connects this line with NLKA and uses component-wise attribution to inform \emph{where} and \emph{how} to steer.

\section{Experimental Setup}

\paragraph{Models.}
We evaluate on GPT2-XL \citep{radford2019language} and LLaMA2-7B \citep{touvron2023llama2openfoundation}, two widely used LLMs in knowledge editing and mechanistic analysis \citep{meng2023masseditingmemorytransformer, scialanga2025sakesteeringactivationsknowledge, gupta2024unifiedframeworkmodelediting}.

\paragraph{Metrics.}
On CounterFact, we report accuracy (Acc; edit success on the rewritten fact), 
generalization (Gen; success on paraphrased prompts), and specificity (Spec; 
preservation of unrelated facts), following standard knowledge editing 
evaluation protocols \citep{meng2022locating}. Following prior KE work 
\citep{scialanga2025sakesteeringactivationsknowledge}, we additionally 
include doubt robustness metrics (DI, DII), originally introduced by 
\citet{Ma2024OnTR}, which evaluate robustness under prompts that introduce 
uncertainty or challenge the edited fact, allowing us to assess the 
contextual robustness of edits.

On Popular, we report Subject Aliasing (SA), which evaluates generalization 
over subject synonyms; Compositionality I (CI) and Compositionality II (CII), 
which measure multi-hop reasoning; and Relation Specificity (RS), which 
measures logical locality by verifying that edits do not alter outputs for 
the same subject under different relations \citep{cohen2023evaluatingrippleeffectsknowledge}. 
Together, these metrics evaluate whether edits generalize consistently to 
logical implications of the modified fact. Detailed metric definitions are provided in App.~\ref{app:metrics}.

\paragraph{Datasets.}
For factual rewriting, we use the CounterFact subset provided by EasyEdit 
(1,031 examples) \citep{wang2024easyediteasytouseknowledgeediting}. 
Due to the computational cost of large-scale editing experiments, we 
evaluate on a fixed subset of 120 edits, which provides sufficient 
coverage for reliable comparison while maintaining practical runtime. For compositional generalization, we use the Popular subset of RippleEdits 
\citep{cohen2023evaluatingrippleeffectsknowledge}. We evaluate on a fixed 
balanced subset of 120 edits (App.~\ref{ripple_dataset})

\section{Mechanistic Attribution of Edited Facts}
\label{sec:mechanistic-attribution}

We use NLKA (§\ref{sec:background}) to quantify how edits redistribute component-wise support for the object token. For each post-edit model produced by the evaluated KE methods (FT, ROME, MEMIT, and IKE), we compute contribution scores $C_{\ell,c}(y)$ per layer $\ell$, component $c \in \{\text{attn},\text{ffn}\}$, and token $y$, and analyze their changes relative to the corresponding pre-edit model:
\[\Delta C_{\ell,c}(y) \;=\; C^{\text{edit}}_{\ell,c}(y) \;-\; C^{\text{base}}_{\ell,c}(y).\]
We focus on cases where the base model predicts the original object correctly and the editor succeeds in rewriting it, and average $\Delta C_{\ell,c}(y_{\text{tgt}})$ and $\Delta C_{\ell,c}(y_{\text{orig}})$ across those edits (details in App.~\ref{app:metrics-attrib}).

\paragraph{Cross-model patterns.}
Figure~\ref{fig:cf-gpt2-memit} contrasts layer-wise $\Delta C_{\ell,c}$ patterns between successful and failed MEMIT edits on GPT2-XL. ROME, FT, and IKE exhibit qualitatively similar trends (App.~\ref{app:cf-gpt2-attrib}), and we observe comparable structures on LLaMA2-7B (App.~\ref{app:cf-llama2-attrib}). Across models and editors, three consistent effects emerge: (i) mid-to-late attention layers carry most of the positive shift toward the new target; (ii) attention and FFNs jointly suppress the original fact in mid layers; and (iii) late residual streams act as high-leverage aggregation points, as suggested by attribution trends. A notable difference is that FFNs contribute minimally to target promotion in GPT2-XL but provide modest positive support in LLaMA2-7B, although attention remains the dominant driver in both. Overall, the shared pattern suggests that edited facts are routed through broadly similar circuit motifs across LLMs, motivating interventions that target these regions.

\section{Mechanism-Guided Multi-Layer Activation Steering}
\label{sec:method}

Our method is inspired by SAKE \citep{scialanga2025sakesteeringactivationsknowledge}, an inference-time activation steering approach that edits behavior without modifying weights. SAKE learns an affine transport map between source and target hidden-state distributions and applies it to the final residual stream via forward hooks gated by a scope detector. Given source activations $h_s$ and target activations $h_t$, it fits $m(h)=Ah+b$ to align the distributions. While SAKE shows that latent distribution matching enables knowledge editing, it treats the residual stream as a monolithic target and intervenes only at the final layer. Guided by the attribution patterns in §\ref{sec:mechanistic-attribution}, we instead introduce mechanism-guided steering targeting attribution-identified control regions instead of heuristic last-layer edits.

\paragraph{What to steer.}
Post-edit attribution shows that target-promoting contributions are concentrated in mid-to-late attention layers and are integrated into the residual stream. This motivates steering along attention-residual pathways where attention outputs are incorporated into the residual state. We therefore use attention-residual steering as the primary intervention mechanism, with attention-only and residual-only variants as ablations.

\paragraph{How to steer.}
For a chosen layer range $L$, we collect activations from source and target prompts, project them into a low-dimensional PCA space (512 dimensions by default), and learn a transport map between the two activation distributions. The learned transformation is applied at inference time via forward hooks on the attention-residual stream for layers $\ell \in L$, gated by the same scope detector used in SAKE. Outside the detected scope, the base model remains unchanged.

\paragraph{Where to intervene in depth.}
Attribution analysis indicates that target-promoting contributions are distributed across mid-to-late attention layers rather than localized at a single depth. Motivated by this pattern, we steer over attribution-aligned mid-to-late windows instead of fixing the final layer. Attribution trends suggest that late layers act as aggregation points. However, windowed mid-to-late steering better reflects the distributed structure revealed by attribution and yields more consistent performance. Our best configurations steer attention-residual representations over attribution-aligned mid-to-late windows, with the optimal range varying by dataset and model (e.g., 15--40 for CounterFact GPT2-XL, 31--47 for Popular GPT2-XL, and 13--31 for LLaMA2-7B; Tables~\ref{tab:cf_best_of}, \ref{tab:pop_best_of}).

We fix the PCA dimension to 512 across models and datasets, which provides a strong trade-off between performance and efficiency. In preliminary experiments, PCA-based steering was more stable and efficient than operating in the full space, and we therefore use it in our reported results.

Overall, our approach preserves the no-weight-update property of activation steering while aligning both the intervention target and depth with mechanistic attribution, replacing heuristic last-layer edits with attribution-guided attention-residual steering.

\section*{Conclusion}

We combined NLKA with activation steering to study how edited facts are implemented inside GPT2-XL and LLaMA2-7B. Across four editors, we found a consistent  pattern: mid-to-late attention promotes the new target while late residual blocks appear to aggregate belief shifts into a decisive preference. Guided by these insights, we introduce a mechanism-guided activation steering approach that targets attribution-aligned attention and residual regions, achieving strong edits on CounterFact and Popular.

\section*{Limitations}

Our experiments focus on single-fact edits in GPT2-XL and LLaMA2-7B. While these models are widely used in knowledge-editing research, it remains unclear whether the same attribution patterns hold in larger or more recent architectures. In addition, we primarily evaluate single-edit scenarios; extending mechanism-guided steering to large-scale multi-edit settings remains an open challenge.

\section*{Acknowledgments}

\bibliography{custom}

\appendix

\section{Dataset Construction}
\label{app:data}

\subsection{CounterFact Subset}

For factual rewriting experiments, we use the CounterFact dataset introduced by \citet{meng2023locatingeditingfactualassociations}, which contains knowledge-editing cases of the form $(e,r,o)\rightarrow(e,r,o^{*})$. We use the EasyEdit implementation of this dataset \citep{wang2024easyediteasytouseknowledgeediting}, which provides a curated version with additional evaluation prompts.

Specifically, we use the EasyEdit dataset variant \texttt{counterfact\_portability\_gpt4.json}, which includes portability prompts designed to test whether edited knowledge transfers to related reasoning tasks. The dataset contains 1,031 editing cases in total. From this set, we select the first 120 edits and evaluate all methods on this fixed subset. These cases include prompts for edit success, paraphrase generalization, and locality tests on unrelated facts.

\subsection{RippleEdits Popular Subset}
\label{ripple_dataset}

To evaluate compositional generalization, we use the Popular subset of RippleEdits \citep{cohen2023evaluatingrippleeffectsknowledge}, which contains 885 edits involving widely known entities. RippleEdits is designed to measure ripple effects of knowledge editing across related facts and reasoning chains.

Because not all RippleEdits cases contain both Compositionality I (CI) and Compositionality II (CII) evaluation prompts, we construct a balanced 120-example subset to ensure sufficient coverage of both metrics. Specifically, we select:
\begin{itemize}[leftmargin=*]
\item 26 edits containing both CI and CII tests,
\item 47 edits containing only CI tests,
\item 47 edits containing only CII tests.
\end{itemize}

This strategy ensures that both compositional metrics are well represented while maintaining a balanced evaluation set. All methods are evaluated on the same fixed subset. For each selected edit, we generate paraphrase prompts following the procedure of \citet{scialanga2025sakesteeringactivationsknowledge}, using the DeepSeek API as the paraphrase generation backend.

\section{Evaluation Metrics}
\label{app:metrics}

\subsection{CounterFact Metrics}

For CounterFact, we report three standard editing metrics. \textbf{Accuracy (Acc)} measures whether the edited model produces the new target for the original rewrite prompt. \textbf{Generalization (Gen)} measures whether the edit transfers to semantically equivalent paraphrases within the intended editing scope. \textbf{Specificity (Spec)} measures locality, i.e., whether the model preserves its original behavior on unrelated prompts outside the editing scope.

Intuitively, a successful edit should change the model's output on in-scope inputs while leaving unrelated inputs unaffected.

\subsection{Doubt Robustness Prompts}
\label{app:doubt-prompts}

We additionally evaluate contextual robustness using the doubt-inducing prompts from \citet{Ma2024OnTR}. Let $p$ denote the original prompt, $o$ the old object, and $o^{*}$ the edited target object.

\paragraph{DI.}
A doubt-inducing prompt of the form:
\begin{quote}
$p$, $o^{*}$ ??? No, it can't be, this contradicts my understanding. It is pretty clear that the answer to $p$ is something else. Rethink and give the correct answer:
\end{quote}

\paragraph{DII.}
A stronger doubt-inducing prompt of the form:
\begin{quote}
$p$, $o^{*}$? Really? But I am pretty sure that [old fact with $o$]. Am I right? Yes, you are right. I apologize for any confusion. Reconsider that the answer to $p$ should be:
\end{quote}

DI and DII measure whether the edited belief remains stable under uncertainty-raising or contradiction-raising contexts.

\subsection{Popular / RippleEdits Metrics}
\label{app:ripple-metrics}

Each edit has the form $(e, r, o) \rightarrow (e, r, o^{*})$, where $e$ is the subject, $r$ the relation, and $o^{*}$ the edited object. For the Popular subset of RippleEdits \citep{cohen2023evaluatingrippleeffectsknowledge}, we evaluate four metrics designed to capture ripple effects of editing.

\paragraph{Compositionality I (CI).}
This metric tests whether the edited fact composes correctly with other facts about the edited object.  
For example, if we edit the sibling of \textit{Prince} to a new person, then the model should also update composed facts such as:
\begin{quote}
Profession of sibling of Prince $\rightarrow$ profession of the new sibling
\end{quote}

\paragraph{Compositionality II (CII).}
This metric tests whether the edit propagates through compositions involving a different subject that refers to the edited entity.  
For example, if the model knows:
\begin{quote}
Founder of Paisley Park Records $\rightarrow$ Prince
\end{quote}
then changing the sibling of \textit{Prince} should also update:
\begin{quote}
Sibling of the founder of Paisley Park Records $\rightarrow$ sibling of Prince after editing
\end{quote}

\paragraph{Subject Aliasing (SA).}
This metric evaluates whether the edit generalizes to aliases of the subject.  
For example, after editing the sibling of \textit{Prince}, the model should produce the same edited result for:
\begin{quote}
Sibling of Prince Roger Nelson $\rightarrow$ same edited sibling
\end{quote}

\paragraph{Relation Specificity (RS).}
This metric measures locality by verifying that relations unrelated to the edited fact remain unchanged.  
For example, editing the sibling of \textit{Prince} should not affect:
\begin{quote}
Mother of Prince
\end{quote}

\section{Neuron-Level Attribution Procedure}
\label{app:metrics-attrib}

We analyze how knowledge edits affect internal computation using neuron-level knowledge attribution (NLKA) \citep{yu2024neuronlevelknowledgeattributionlarge}. The method decomposes the log-probability of a target token into additive contributions from transformer components.

\paragraph{Residual stream decomposition.}
For each layer $\ell$, the residual stream evolves as
\[
\begin{aligned}
\mathrm{Lres}_{\ell} &= \mathrm{Lin}_{\ell} + \mathrm{ATTN}_{\ell},\\
\mathrm{Lout}_{\ell} &= \mathrm{Lres}_{\ell} + \mathrm{FFN}_{\ell}.
\end{aligned}
\]

Here $\mathrm{Lin}_{\ell}$ denotes the input to layer $\ell$, $\mathrm{ATTN}_{\ell}$ and $\mathrm{FFN}_{\ell}$ are the attention and feed-forward outputs at that layer, $\mathrm{Lres}_{\ell}$ is the intermediate residual state after adding attention, and $\mathrm{Lout}_{\ell}$ is the final layer output.

\paragraph{Contribution scores.}
Following \citet{yu2024neuronlevelknowledgeattributionlarge}, the contribution of a component $c$ at layer $\ell$ to a target token $y$ is defined as the change in log-probability when the component output is added to the residual stream:
\[
C_{\ell,c}(y) =
\log p(y \mid h + c_{\ell}) -
\log p(y \mid h),
\]
where $h$ denotes the residual state before the component and $c_{\ell}$ is the component output. 

For each prompt, we compute contribution scores for attention and FFN components across all layers by measuring how their outputs change the log-probability of the target token. Token probabilities are obtained by projecting the residual state through the final layer normalization and LM head. This attribution procedure is purely observational and does not modify the model parameters.

\paragraph{Edit comparison.}
To isolate editing effects, we compare contribution scores between edited and base models:
\[
\Delta C_{\ell,c}(y) =
C^{\text{edit}}_{\ell,c}(y) -
C^{\text{base}}_{\ell,c}(y).
\]

\paragraph{Aggregation.}
We compute $\Delta C_{\ell,c}(y)$ for both the new target token $y_{\text{tgt}}$ and the original token $y_{\text{orig}}$, and average results across all successful edits in the evaluation set.

\section{Additional Attribution Analysis}
\label{app:attrib}

\subsection{CounterFact}

\subsubsection{GPT2-XL}
\label{app:cf-gpt2-attrib}

\begin{figure*}[t]
\centering
\begin{tabular}{cc}
\includegraphics[width=8cm]{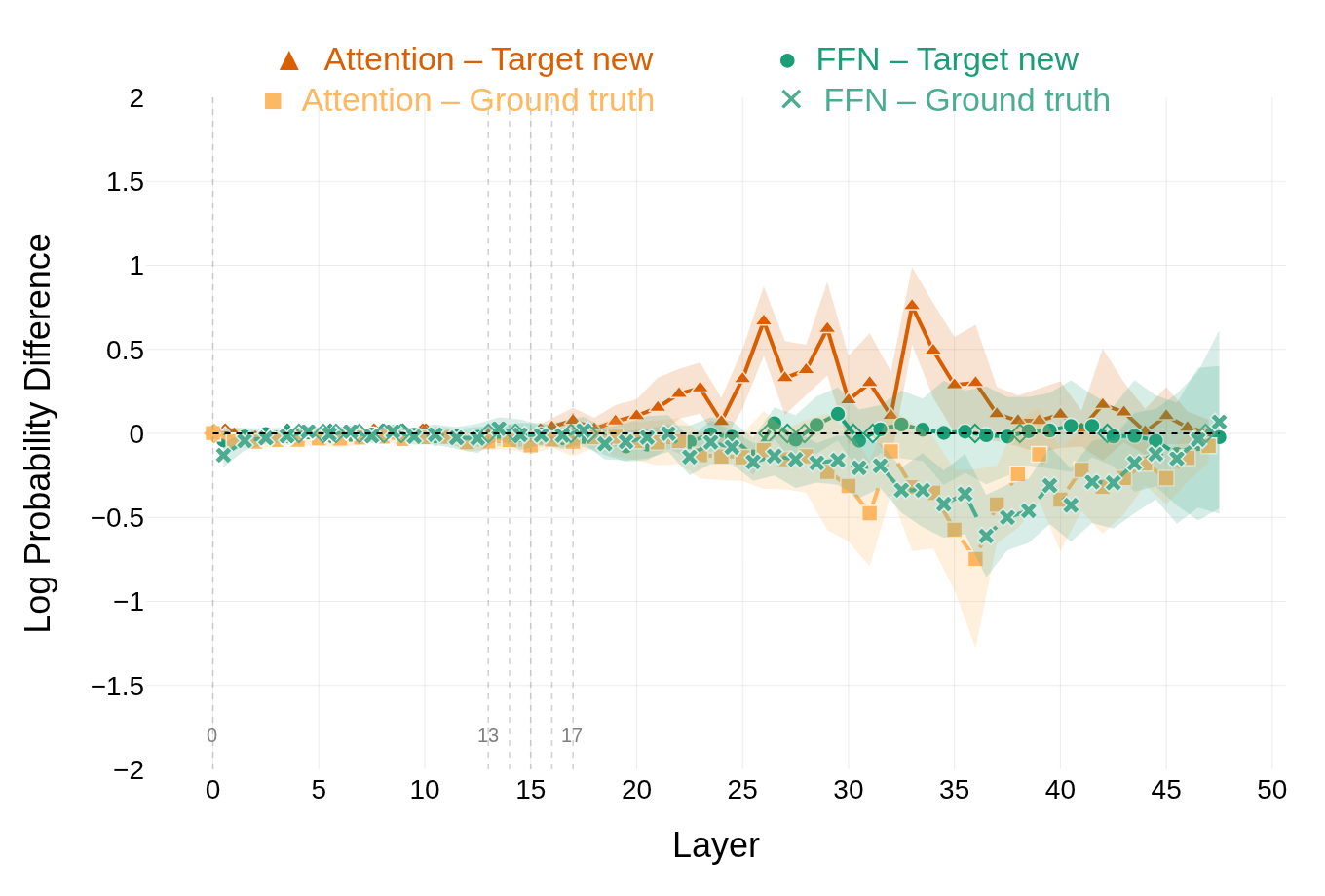} &
\includegraphics[width=8cm]{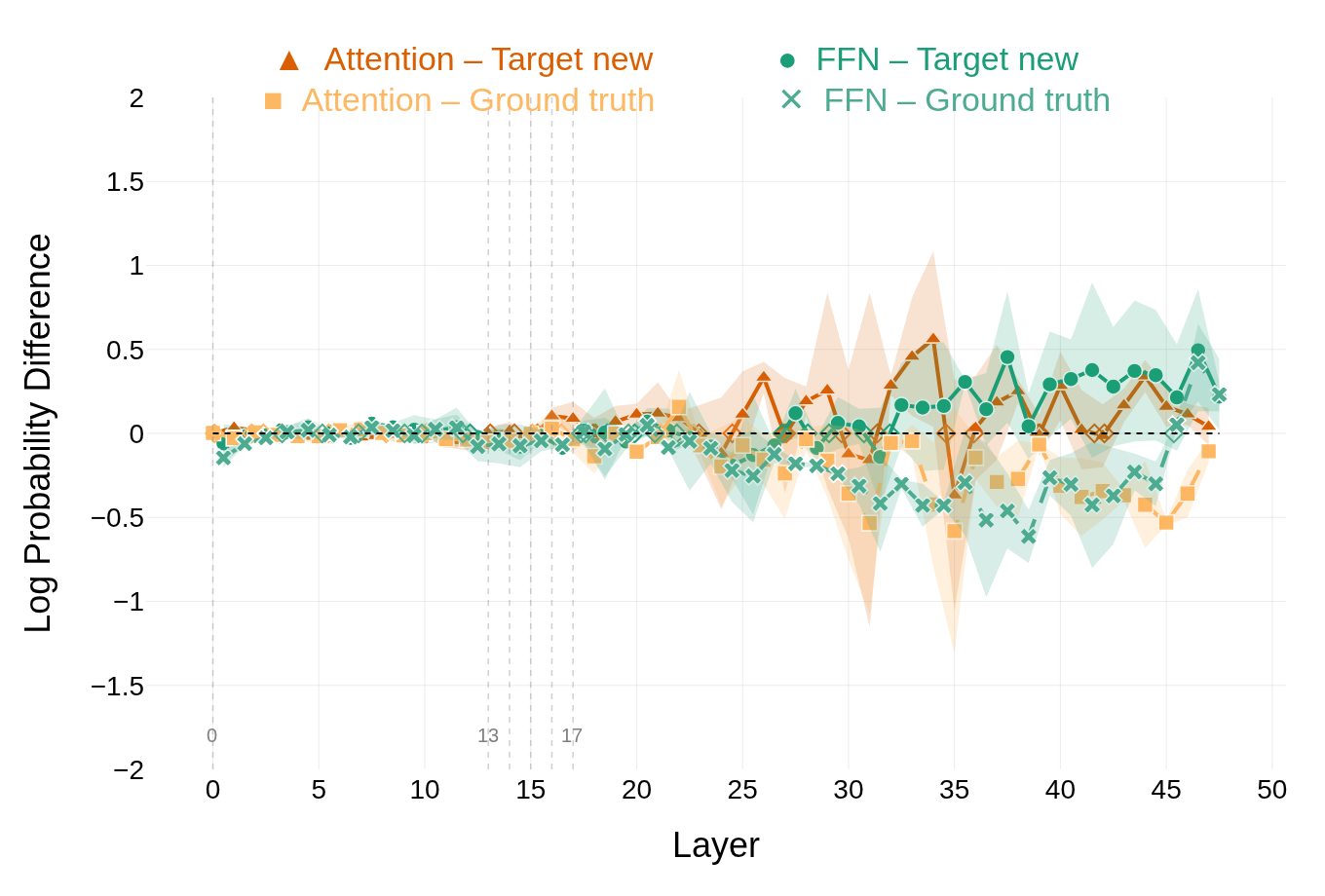} \\
(a) Success (ORI) & (b) Failure (ORI)
\end{tabular}
\caption{\textbf{IKE success vs.\ failure on CounterFact (GPT2-XL).}}
\label{fig:cf-gpt2-ike}
\end{figure*}

\begin{figure*}[h]
    \makebox[\textwidth][c]{%
        \begin{tabular}{cc}
            \includegraphics[width=8.0cm]{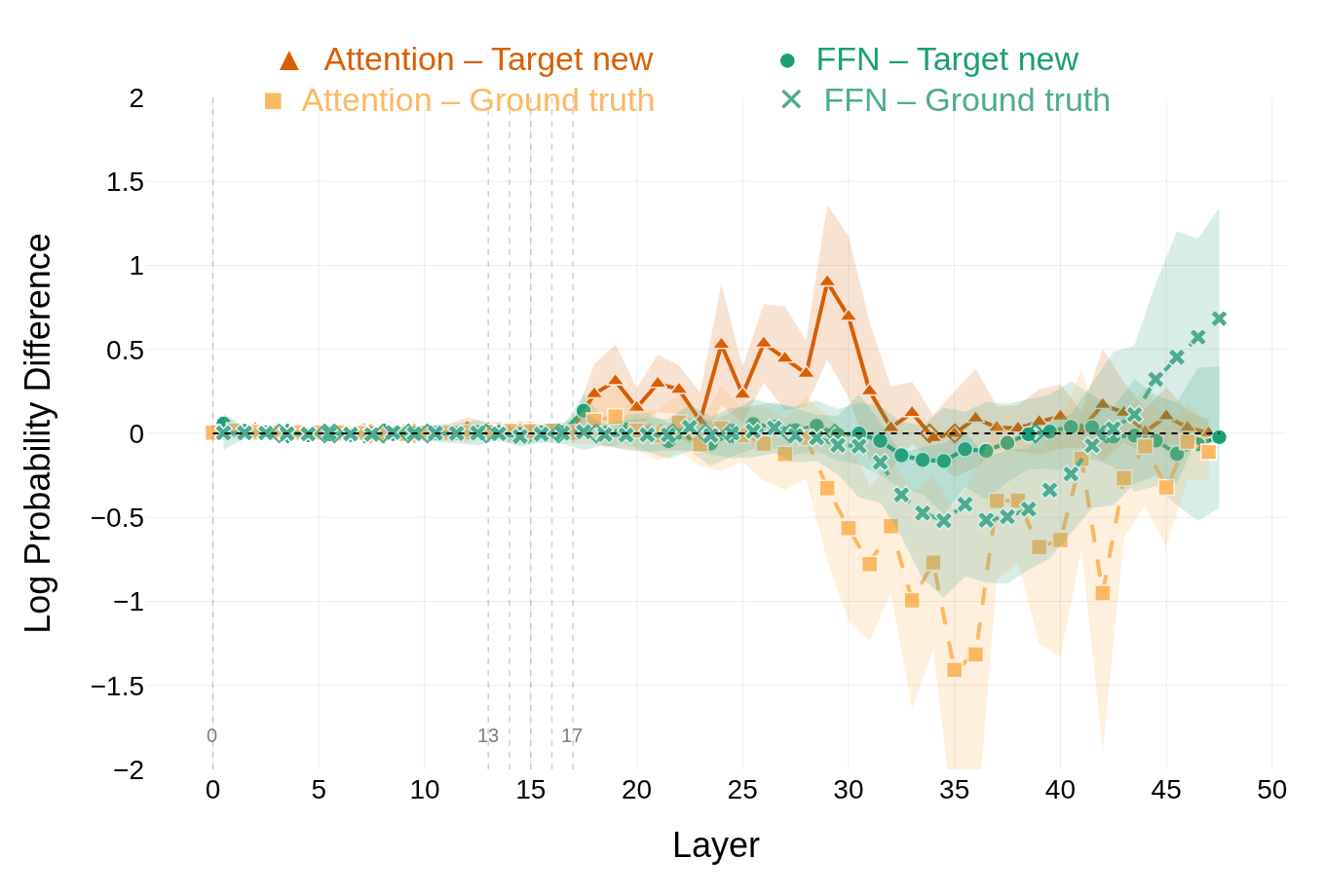} &
            \includegraphics[width=8.0cm]{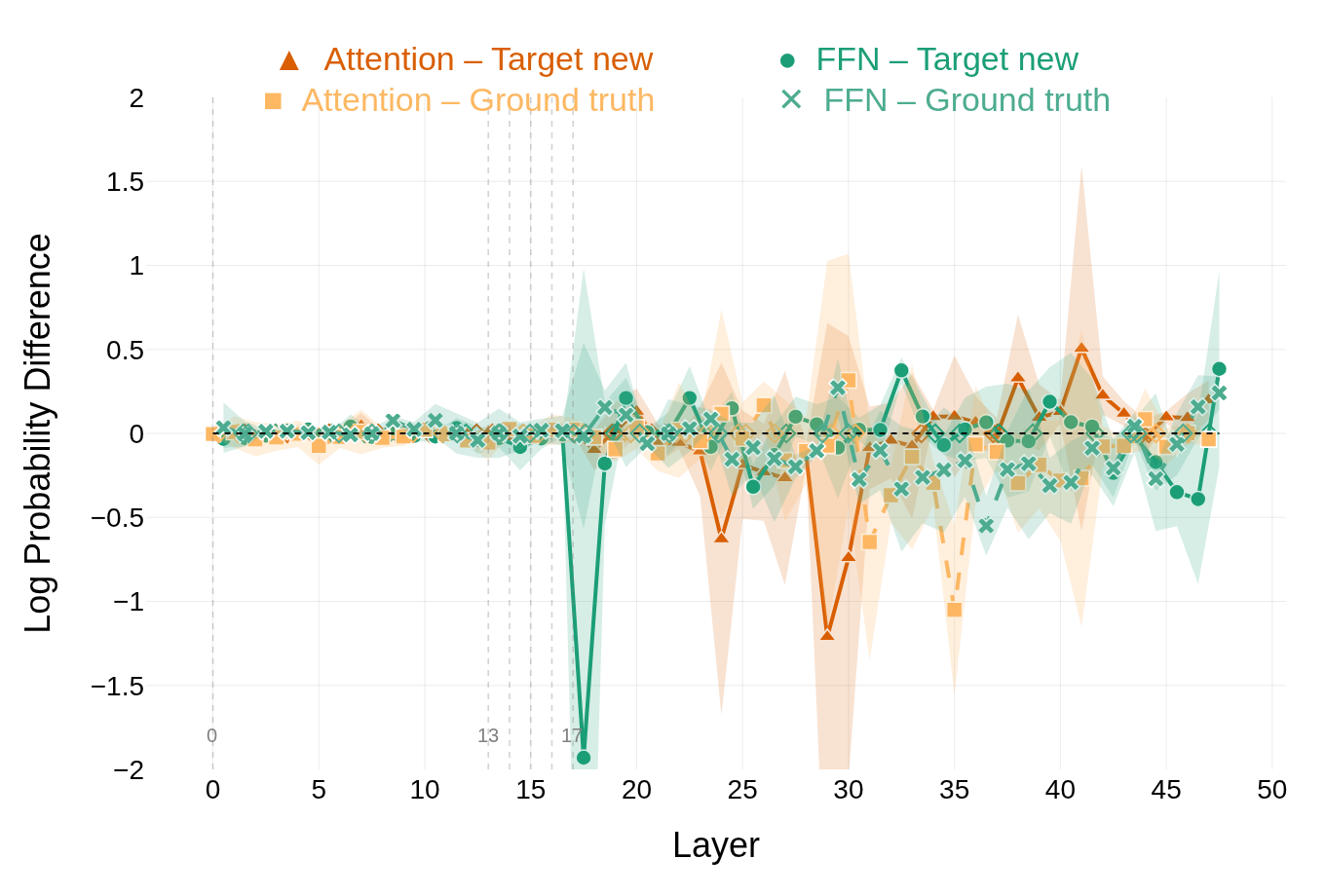} \\
            (a) Success (L17) & (b) Failure (L17) \\
        \end{tabular}%
    }
    \caption{\textbf{ROME success vs.\ failure on CounterFact (GPT2-XL).}}
    \label{fig:cf-gpt2-rome}
\end{figure*}

\begin{figure*}[t]
\centering
\begin{tabular}{cc}
\includegraphics[width=8cm]{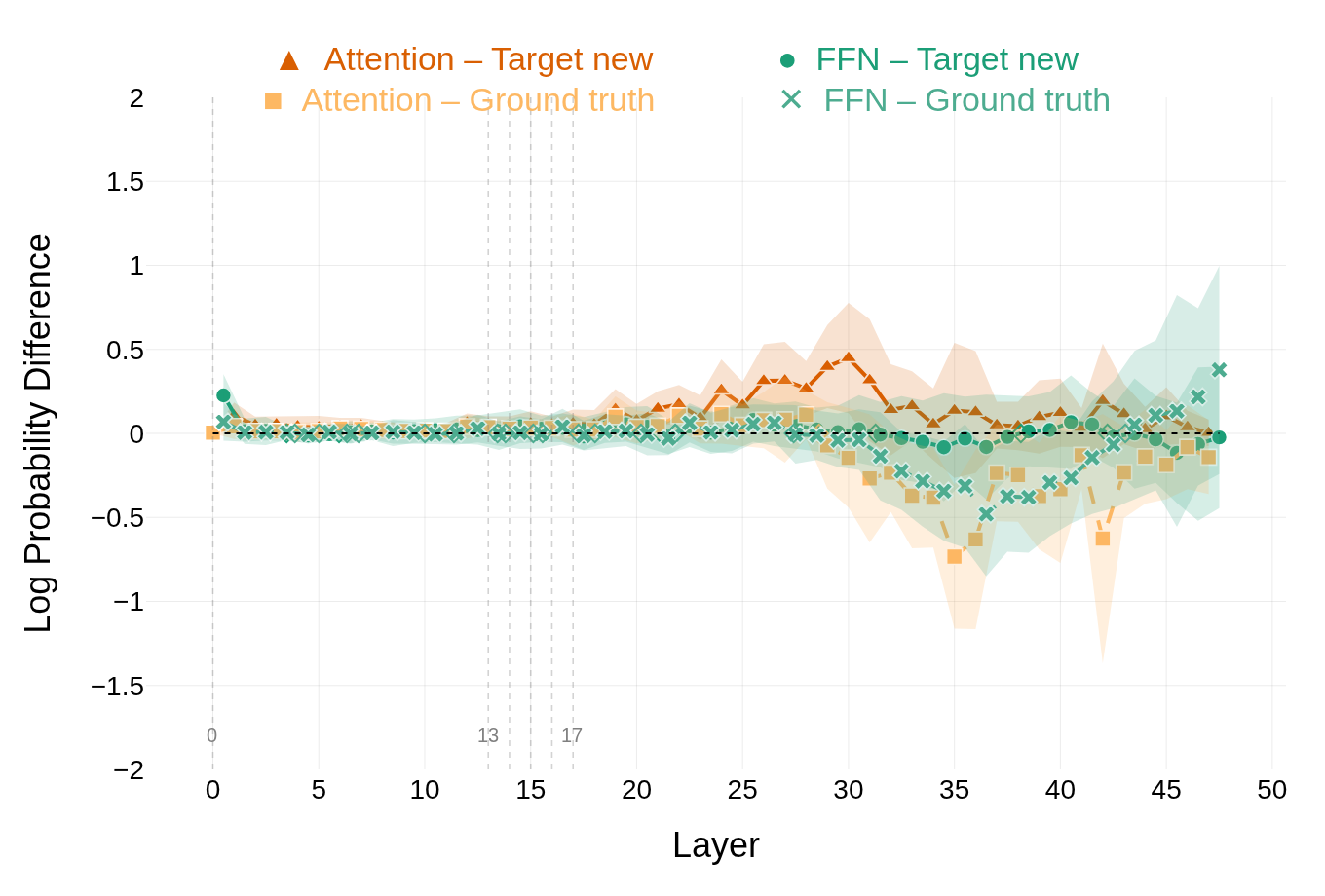} &
\includegraphics[width=8cm]{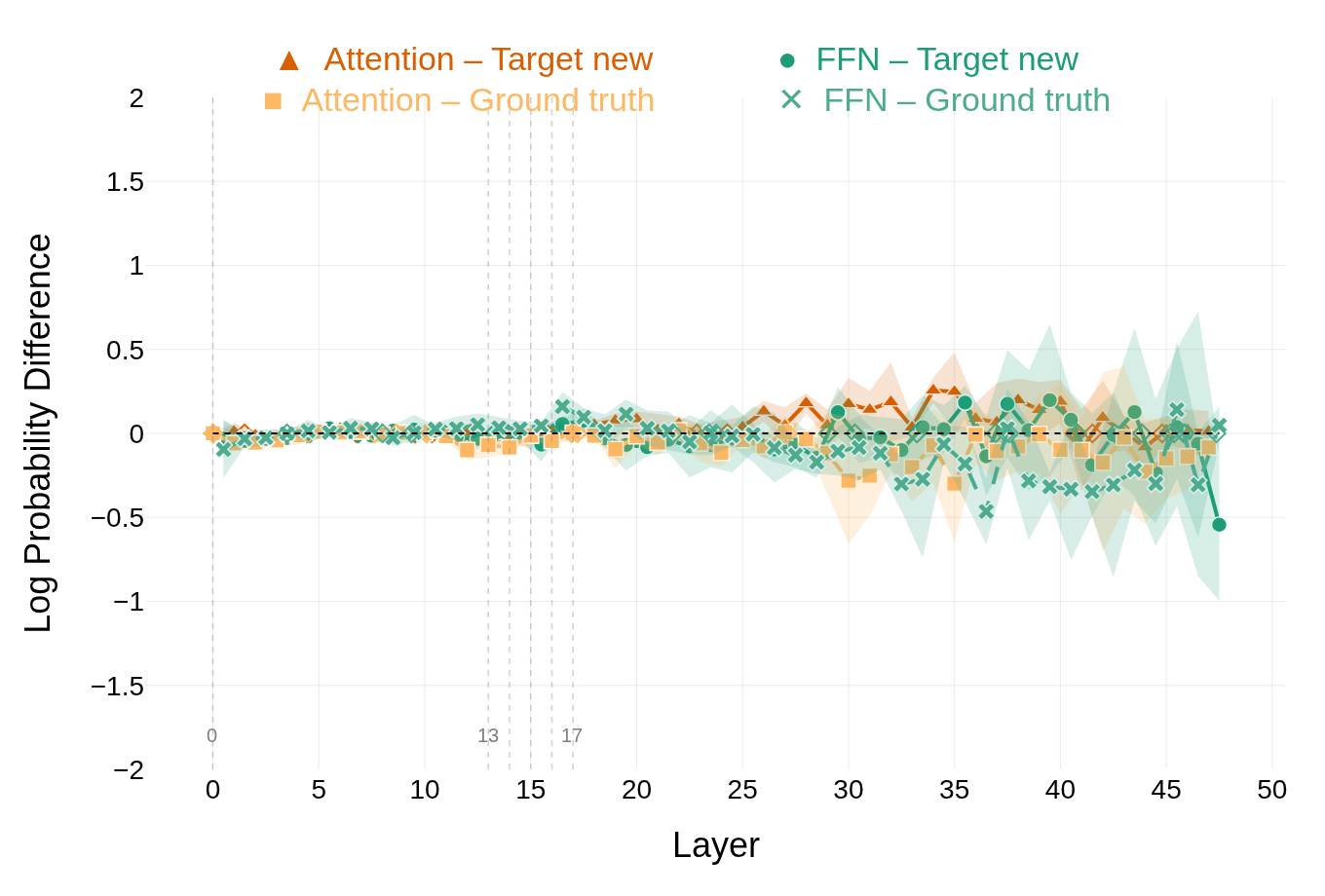} \\
(a) FT Success (L0) & (b) FT Failure (L0)
\end{tabular}
\caption{\textbf{Fine-tuning success vs.\ failure on CounterFact (GPT2-XL).}}
\label{fig:cf-gpt2-ft}
\end{figure*}

Figures~\ref{fig:cf-gpt2-ike}, \ref{fig:cf-gpt2-rome}, and \ref{fig:cf-gpt2-ft}
show attribution patterns for successful and failed edits under IKE, ROME,
and fine-tuning on GPT2-XL for CounterFact.

\subsubsection{LLaMA2-7B}
\label{app:cf-llama2-attrib}

\begin{figure*}[t]
\centering
\begin{tabular}{cc}
\includegraphics[width=8cm]{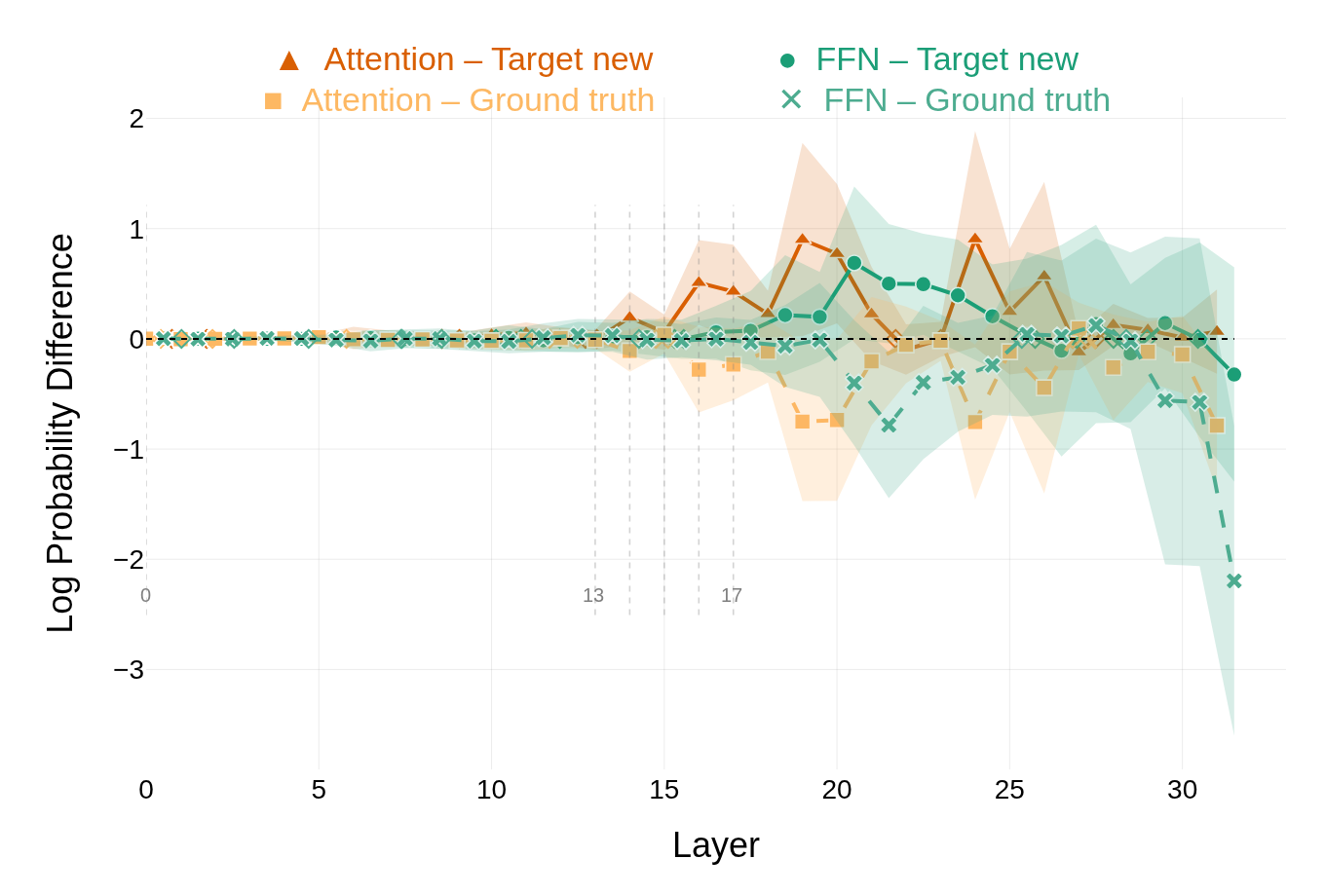} &
\includegraphics[width=8cm]{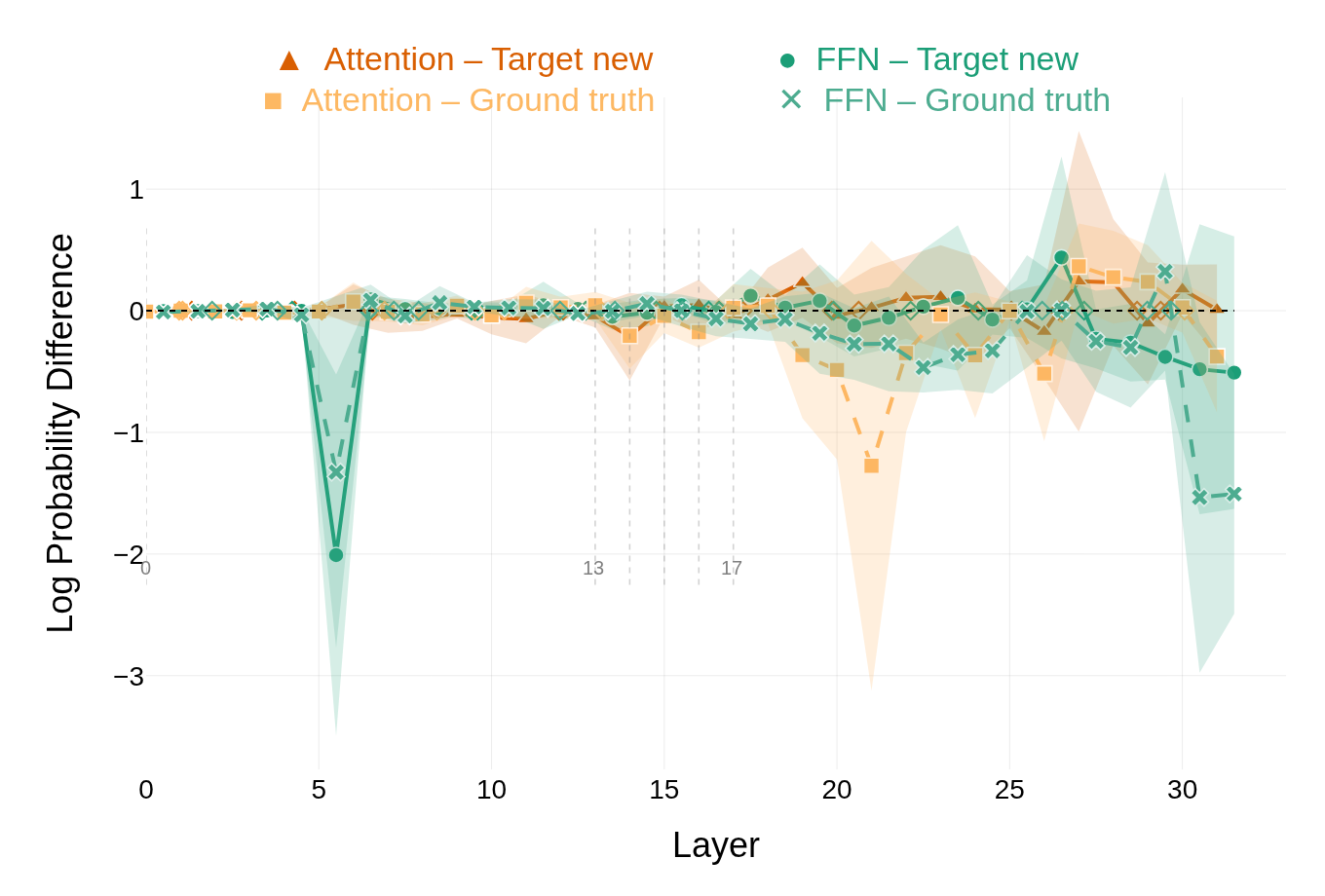} \\
(a) Success (L5) & (b) Failure (L5)
\end{tabular}
\caption{\textbf{ROME success vs.\ failure on CounterFact (LLaMA2-7B).}}
\label{fig:cf-llama-rome}
\end{figure*}

\begin{figure*}[t]
\centering
\begin{tabular}{cc}
\includegraphics[width=8cm]{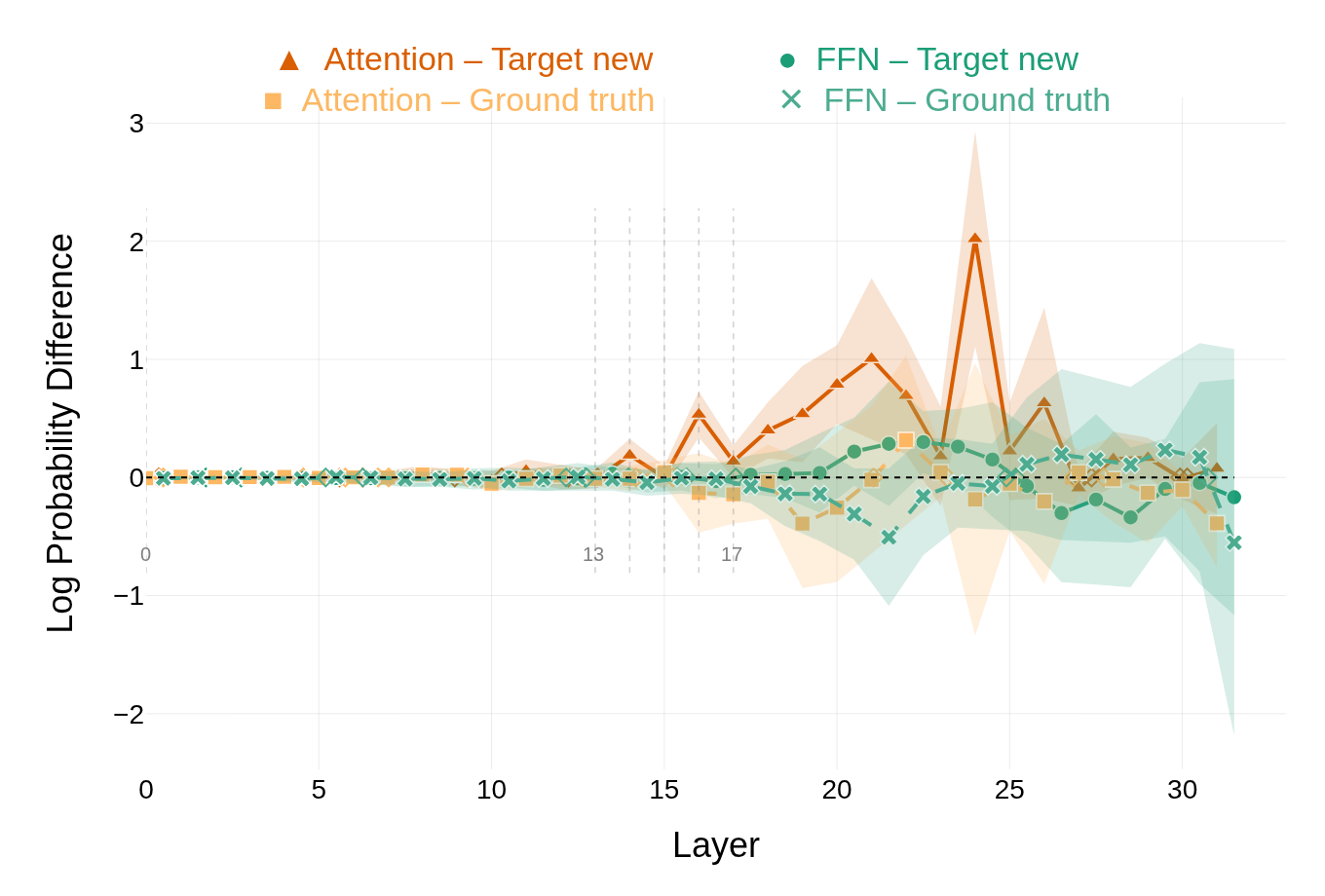} &
\includegraphics[width=8cm]{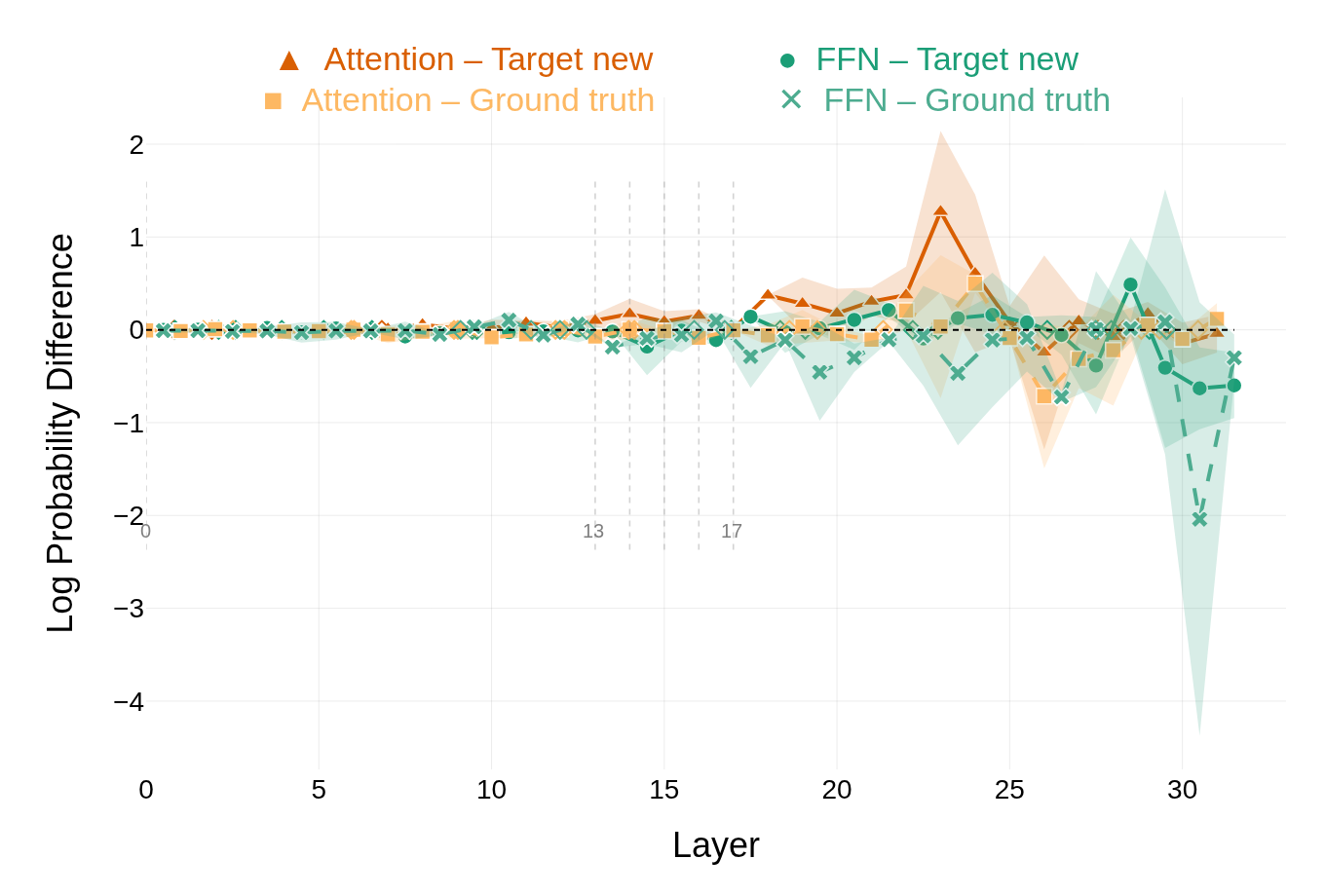} \\
(a) Success (ORI) & (b) Failure (ORI)
\end{tabular}
\caption{\textbf{IKE success vs.\ failure on CounterFact (LLaMA2-7B).}}
\label{fig:cf-llama-ike}
\end{figure*}

\begin{figure*}[t]
\centering
\begin{tabular}{cc}
\includegraphics[width=8cm]{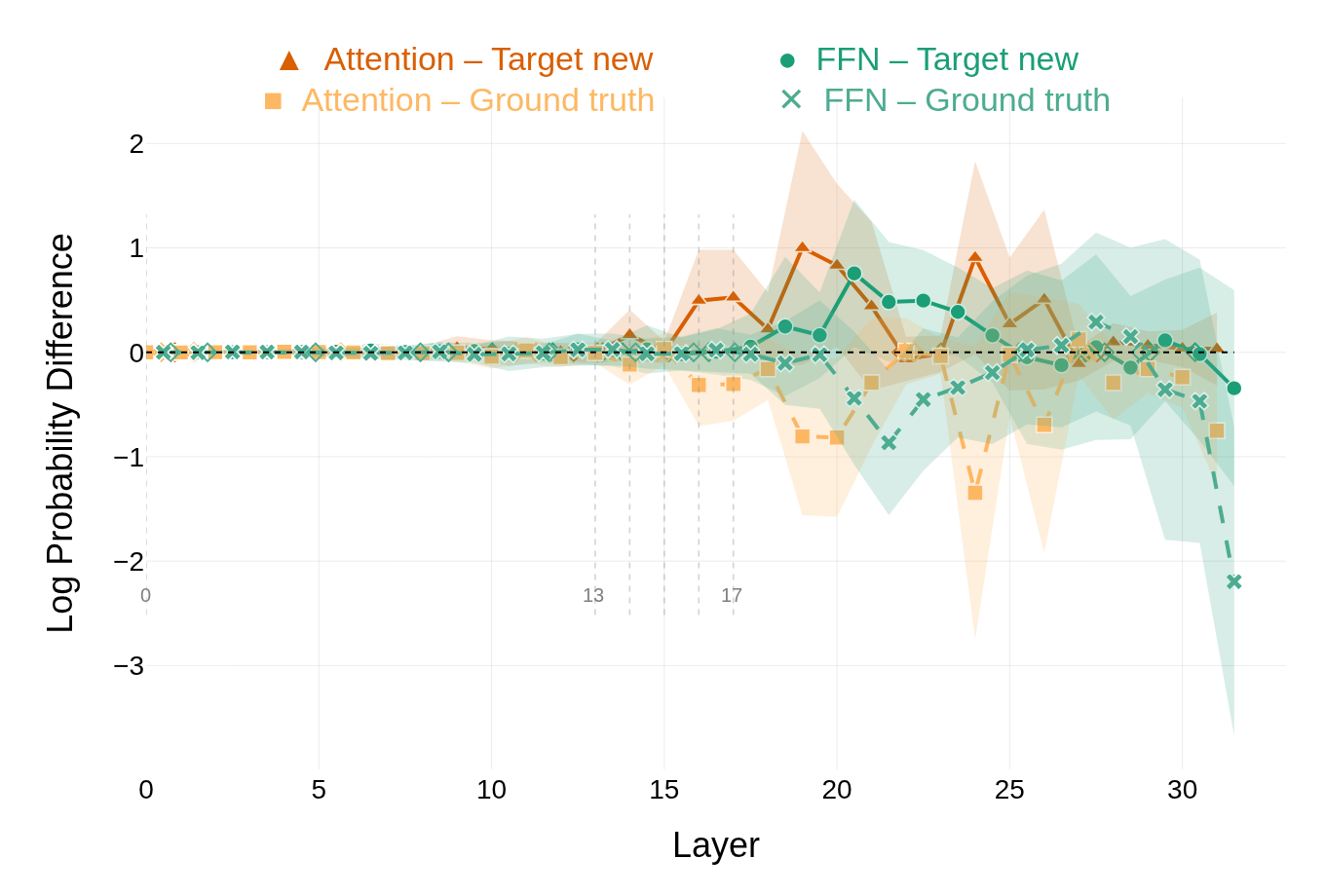} &
\includegraphics[width=8cm]{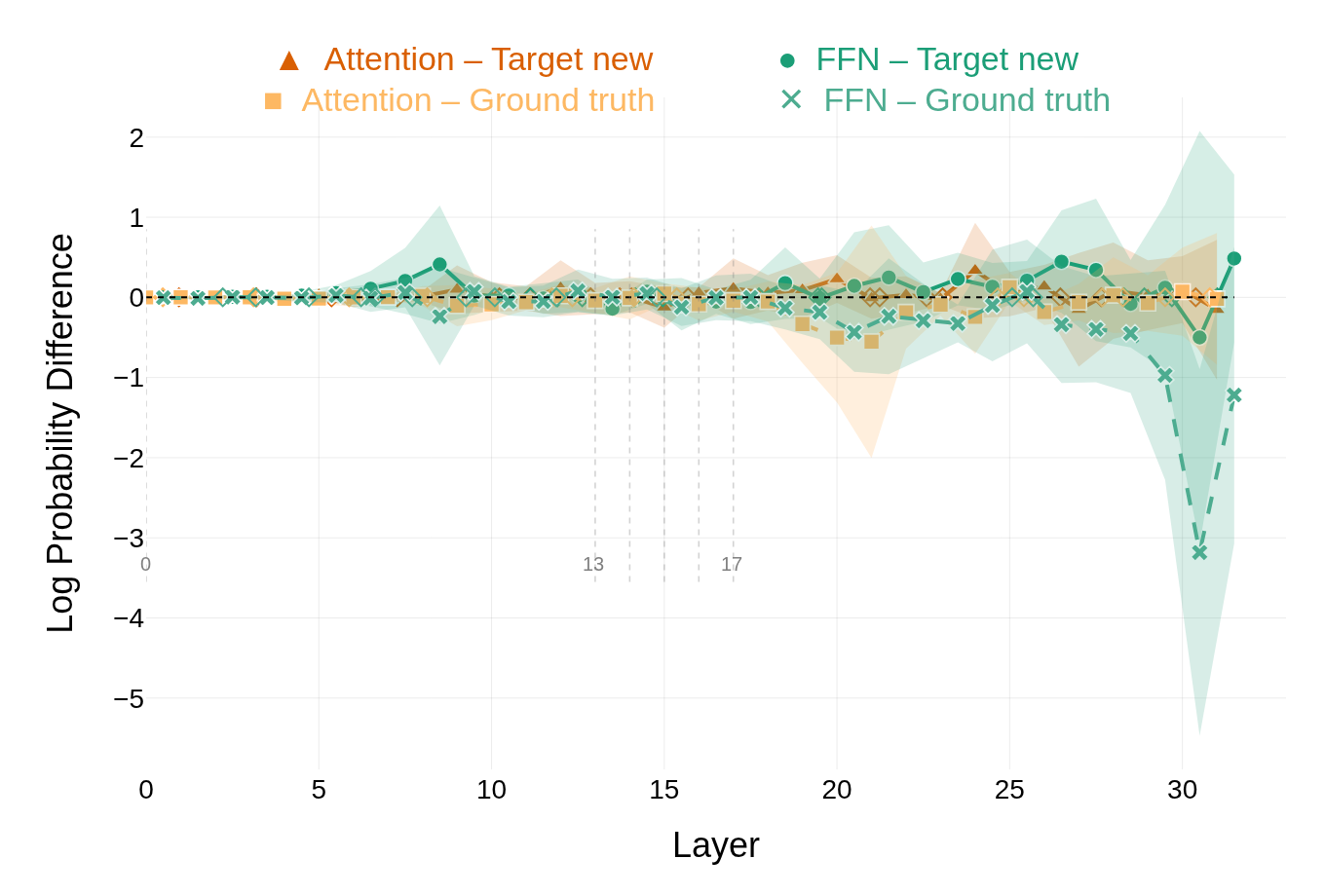} \\
(a) Success (L4--8) & (b) Failure (L4--8)
\end{tabular}
\caption{\textbf{MEMIT success vs.\ failure on CounterFact (LLaMA2-7B).}}
\label{fig:cf-llama-memit}
\end{figure*}

\begin{figure*}[t]
\centering
\begin{tabular}{cc}
\includegraphics[width=8cm]{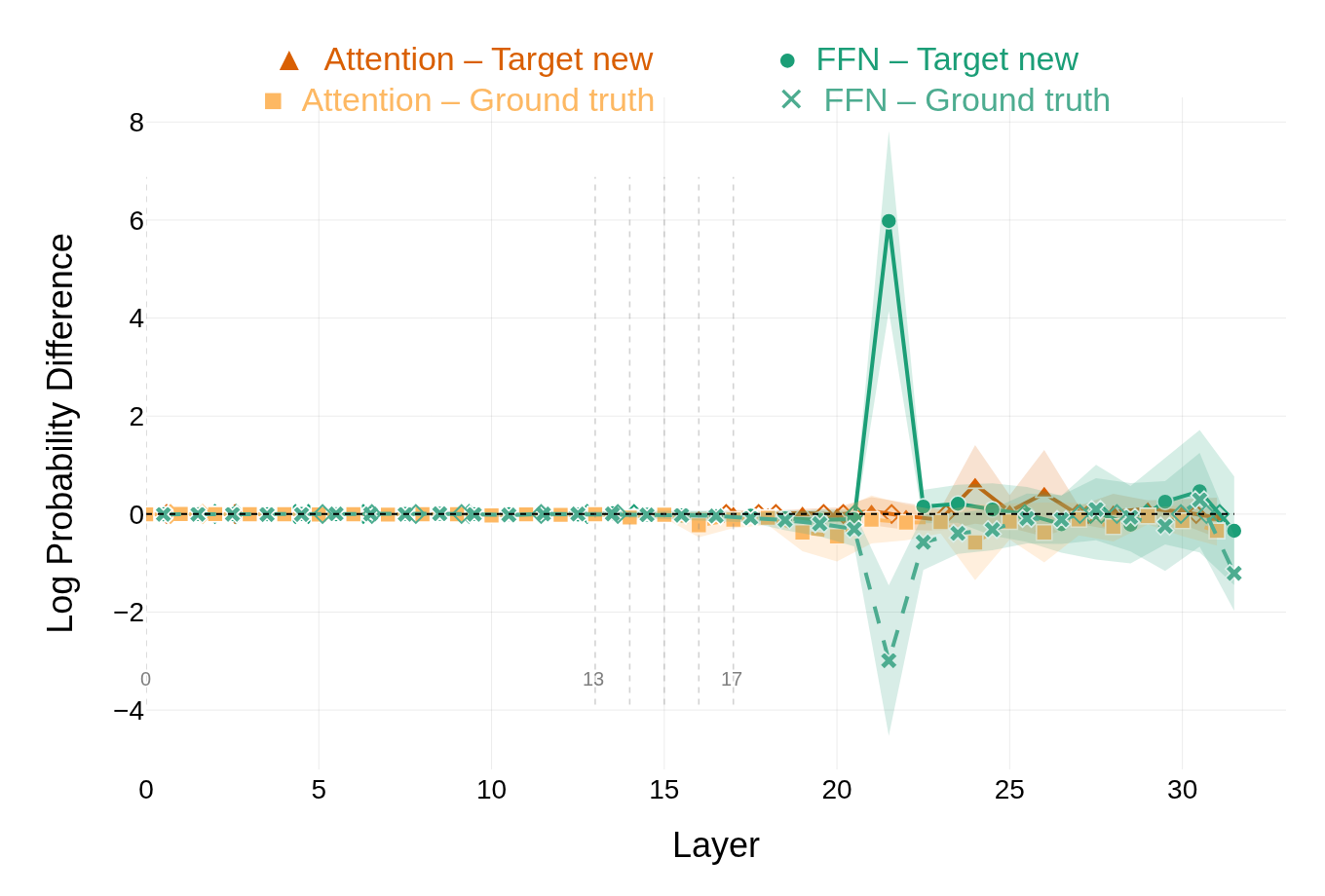} &
\includegraphics[width=8cm]{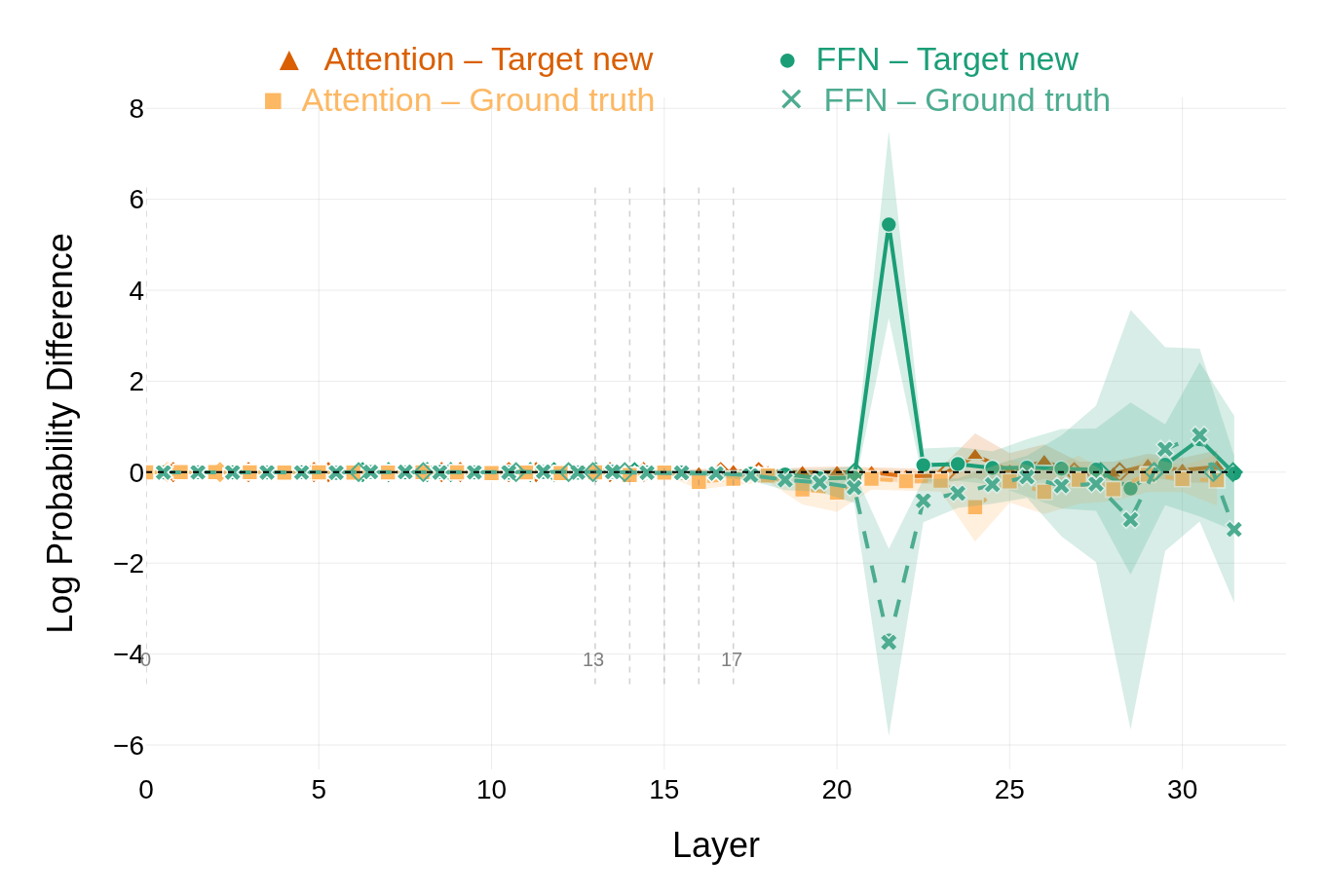} \\
(a) FT Success (L21) & (b) FT Failure (L21)
\end{tabular}
\caption{\textbf{Fine-tuning success vs.\ failure on CounterFact (LLaMA2-7B).}}
\label{fig:cf-llama-ft}
\end{figure*}

Figures~\ref{fig:cf-llama-rome}, \ref{fig:cf-llama-ike},
\ref{fig:cf-llama-memit}, and \ref{fig:cf-llama-ft}
show attribution patterns for successful and failed edits under ROME, IKE,
MEMIT, and fine-tuning on LLaMA2-7B for CounterFact.

\subsection{Popular}

\subsubsection{GPT2-XL}

\begin{figure*}[t]
\centering
\begin{tabular}{cc}
\includegraphics[width=8cm]{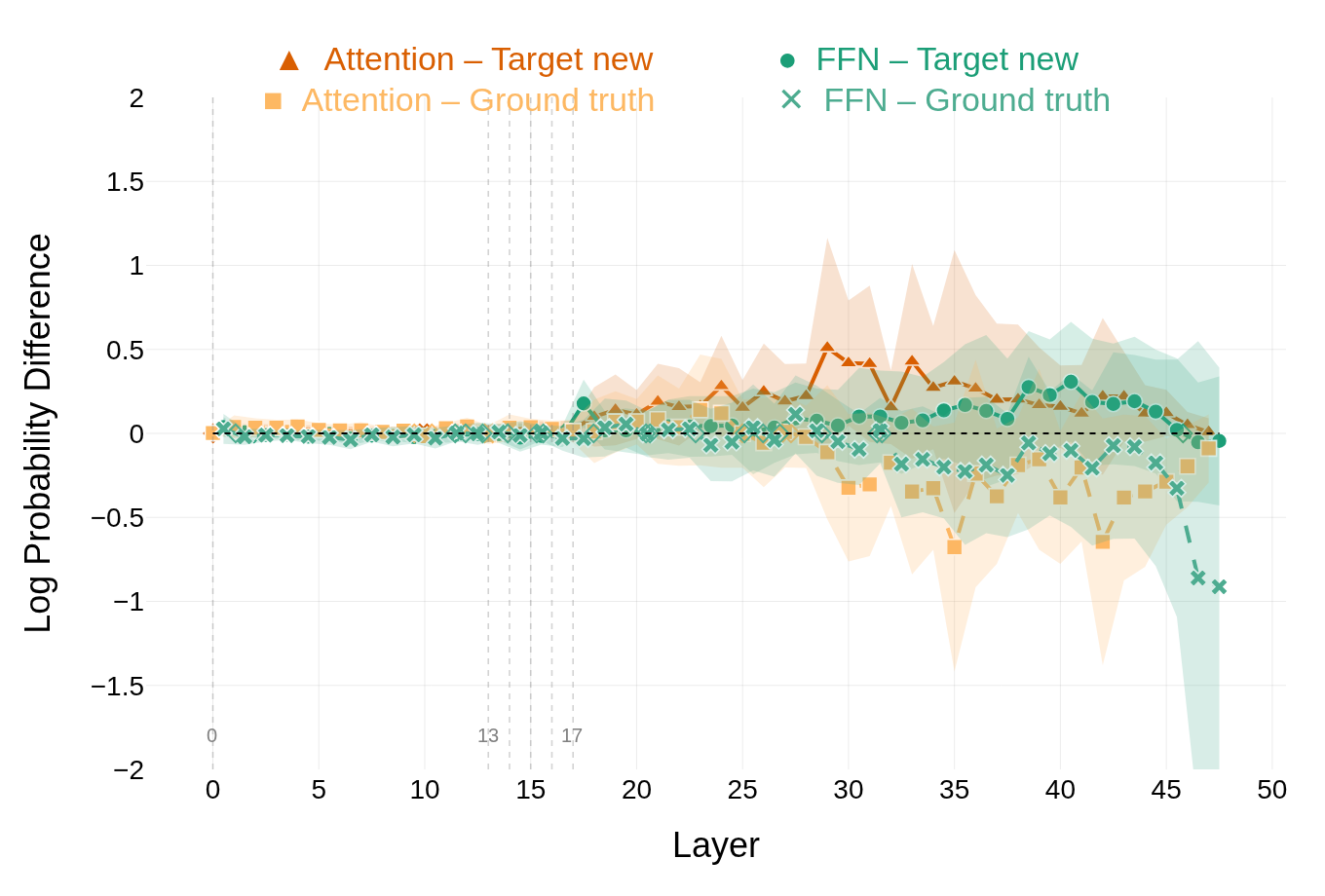} &
\includegraphics[width=8cm]{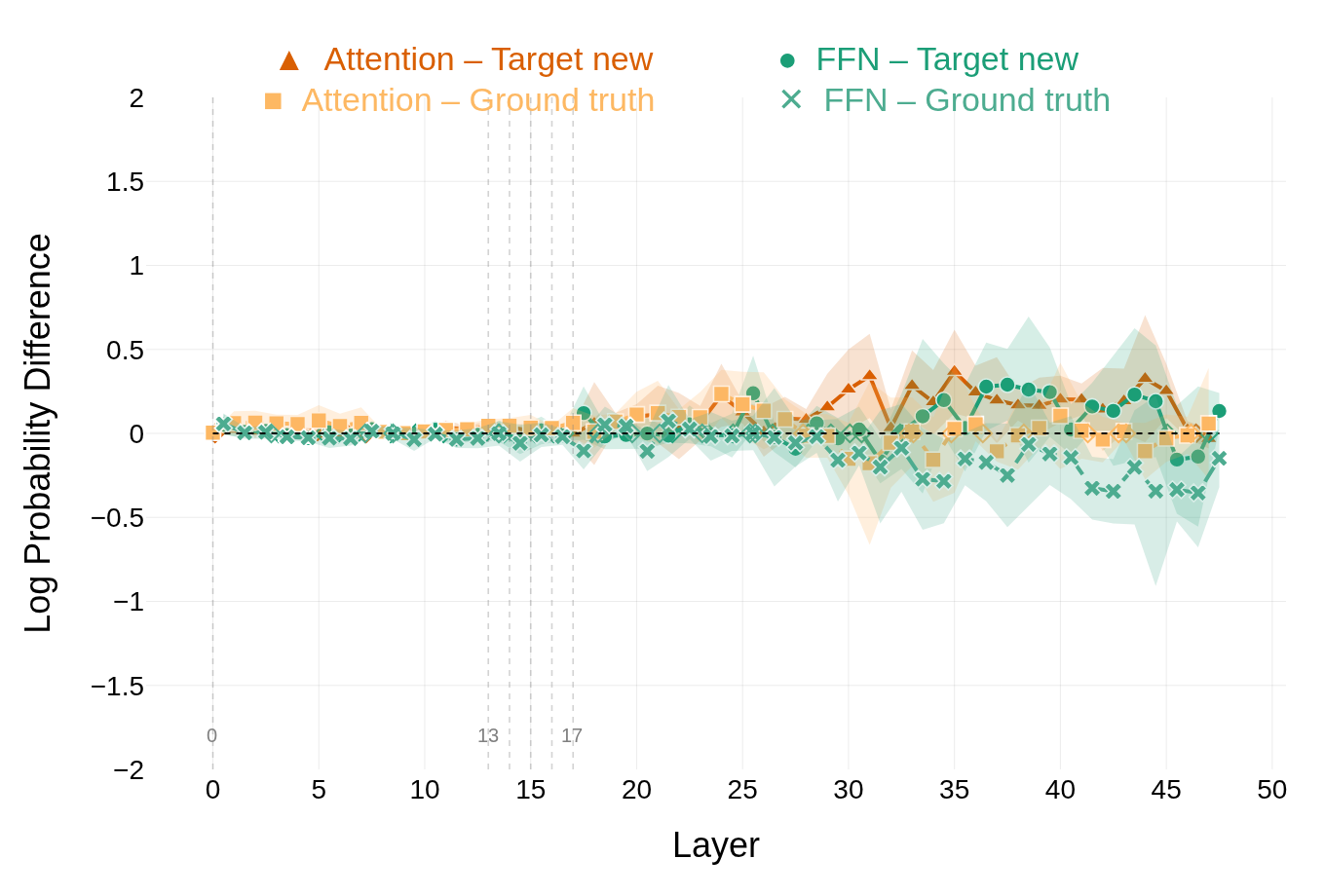} \\
(a) Success (L17) & (b) Failure (L17)
\end{tabular}
\caption{\textbf{ROME success vs.\ failure on Popular (GPT2-XL).}}
\label{fig:pop-gpt2-rome}
\end{figure*}

\begin{figure*}[t]
\centering
\begin{tabular}{cc}
\includegraphics[width=8cm]{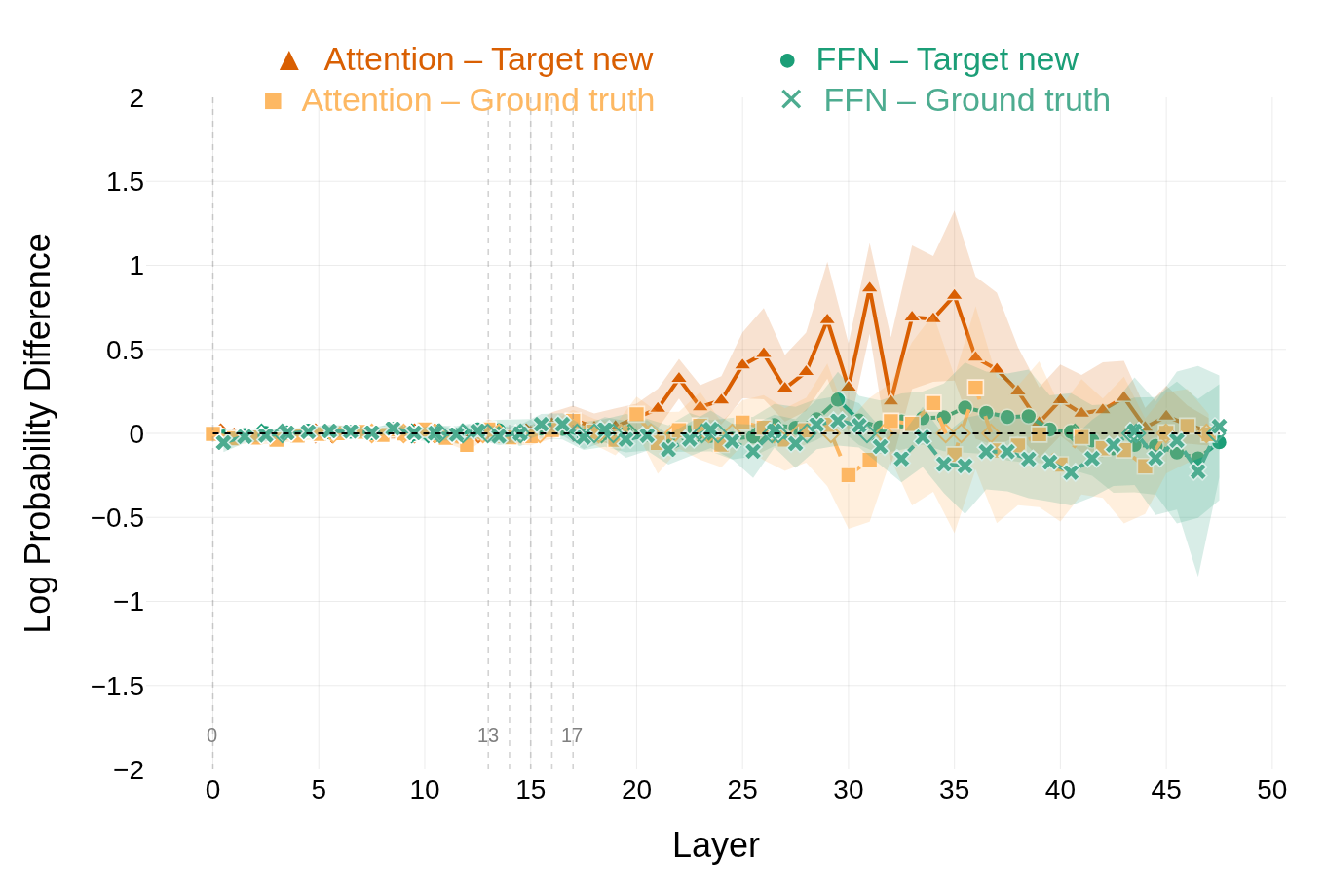} &
\includegraphics[width=8cm]{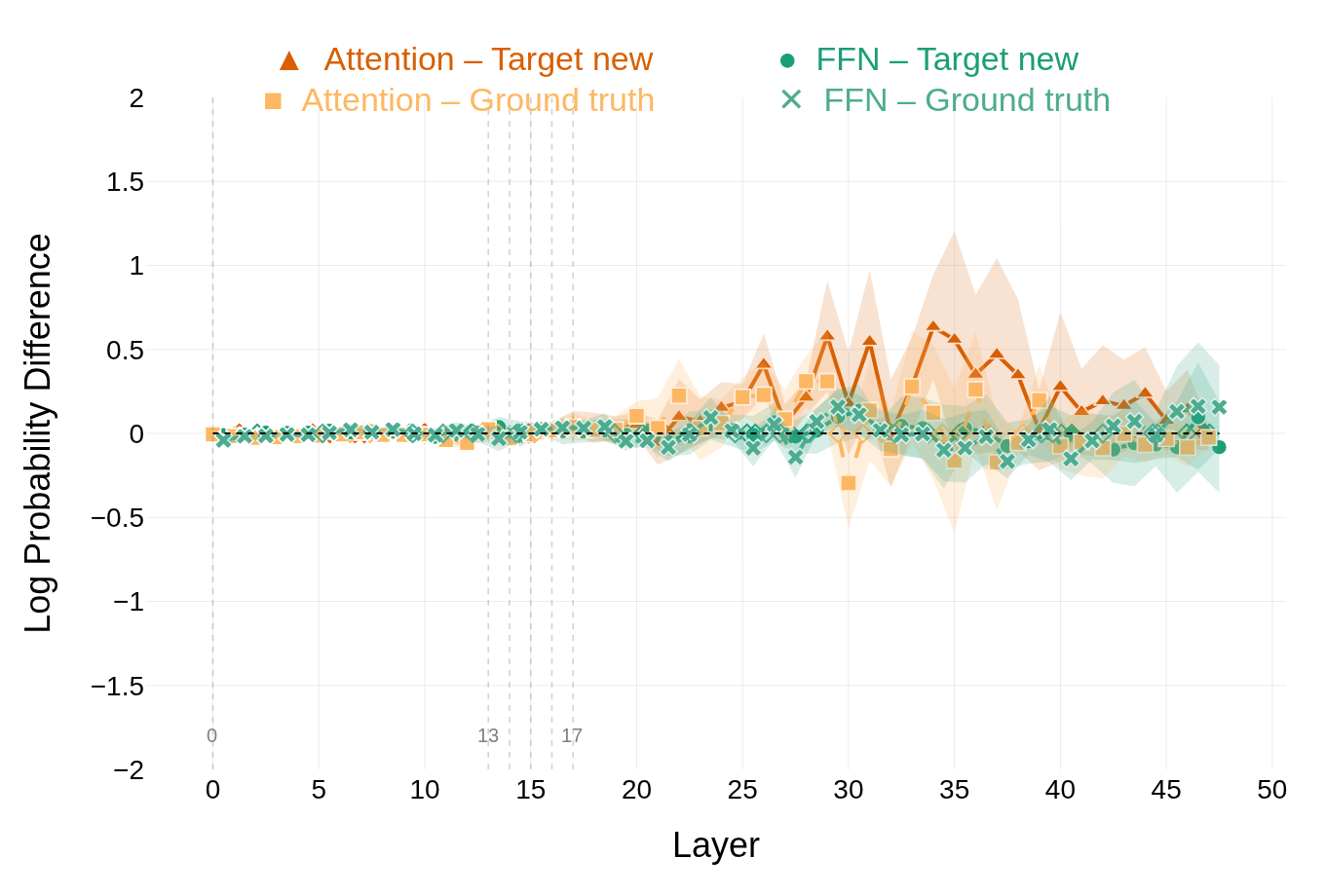} \\
(a) Success (ORI) & (b) Failure (ORI)
\end{tabular}
\caption{\textbf{IKE success vs.\ failure on Popular (GPT2-XL).}}
\label{fig:pop-gpt2-ike}
\end{figure*}

\begin{figure*}[t]
\centering
\begin{tabular}{cc}
\includegraphics[width=8cm]{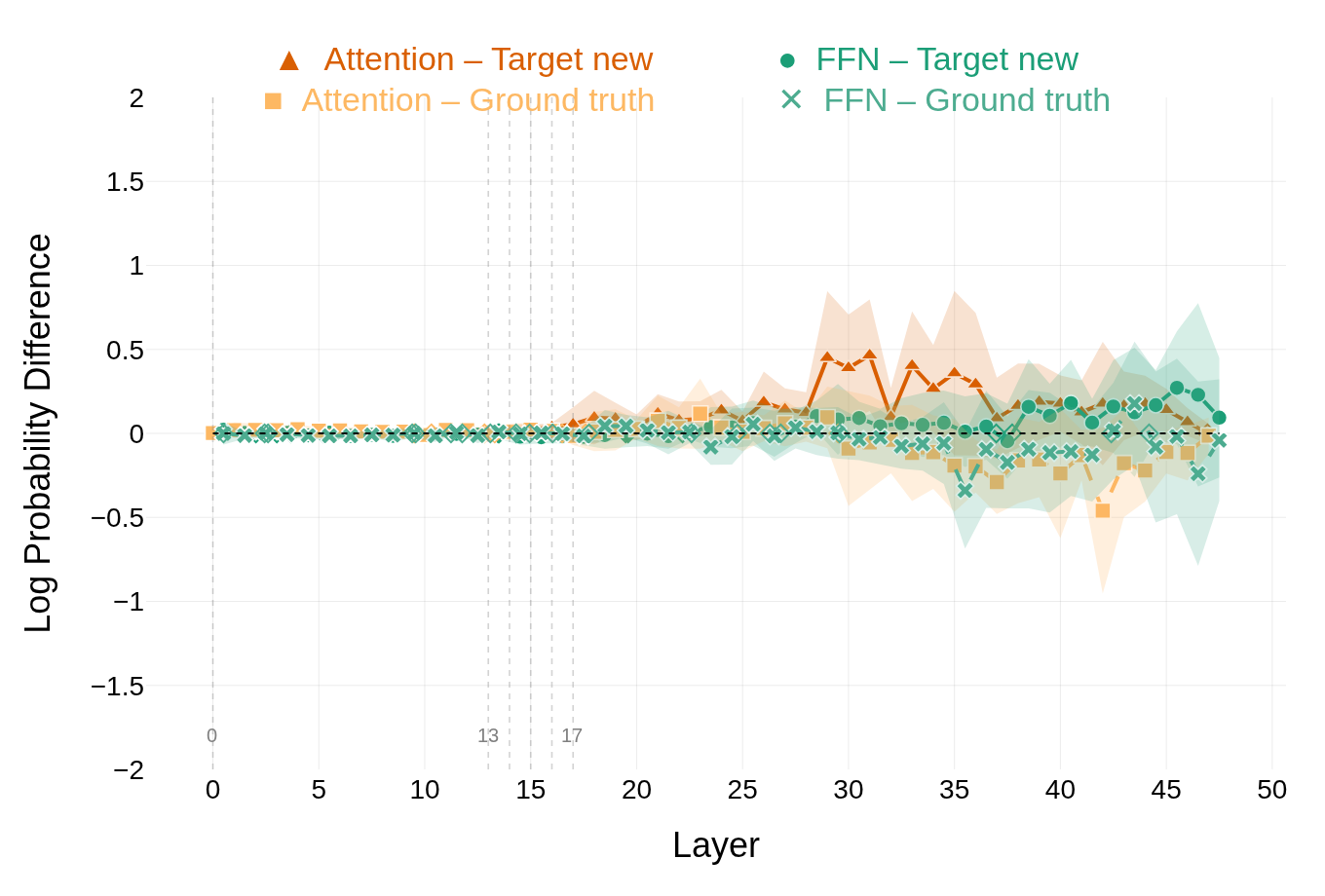} &
\includegraphics[width=8cm]{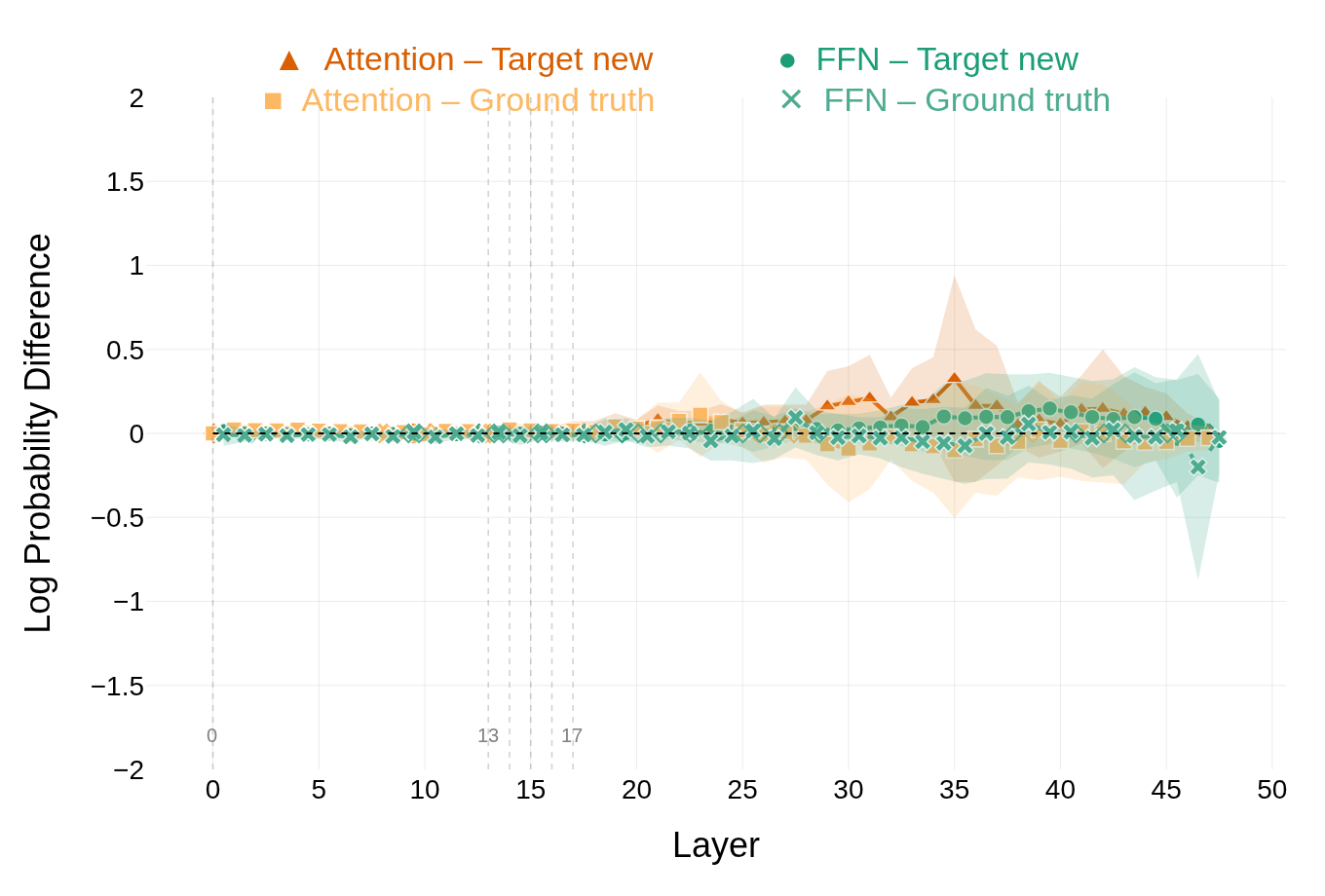} \\
(a) Success (L13--17) & (b) Failure (L13--17)
\end{tabular}
\caption{\textbf{MEMIT success vs.\ failure on Popular (GPT2-XL).}}
\label{fig:pop-gpt2-memit}
\end{figure*}

\begin{figure*}[t]
\centering
\begin{tabular}{cc}
\includegraphics[width=8cm]{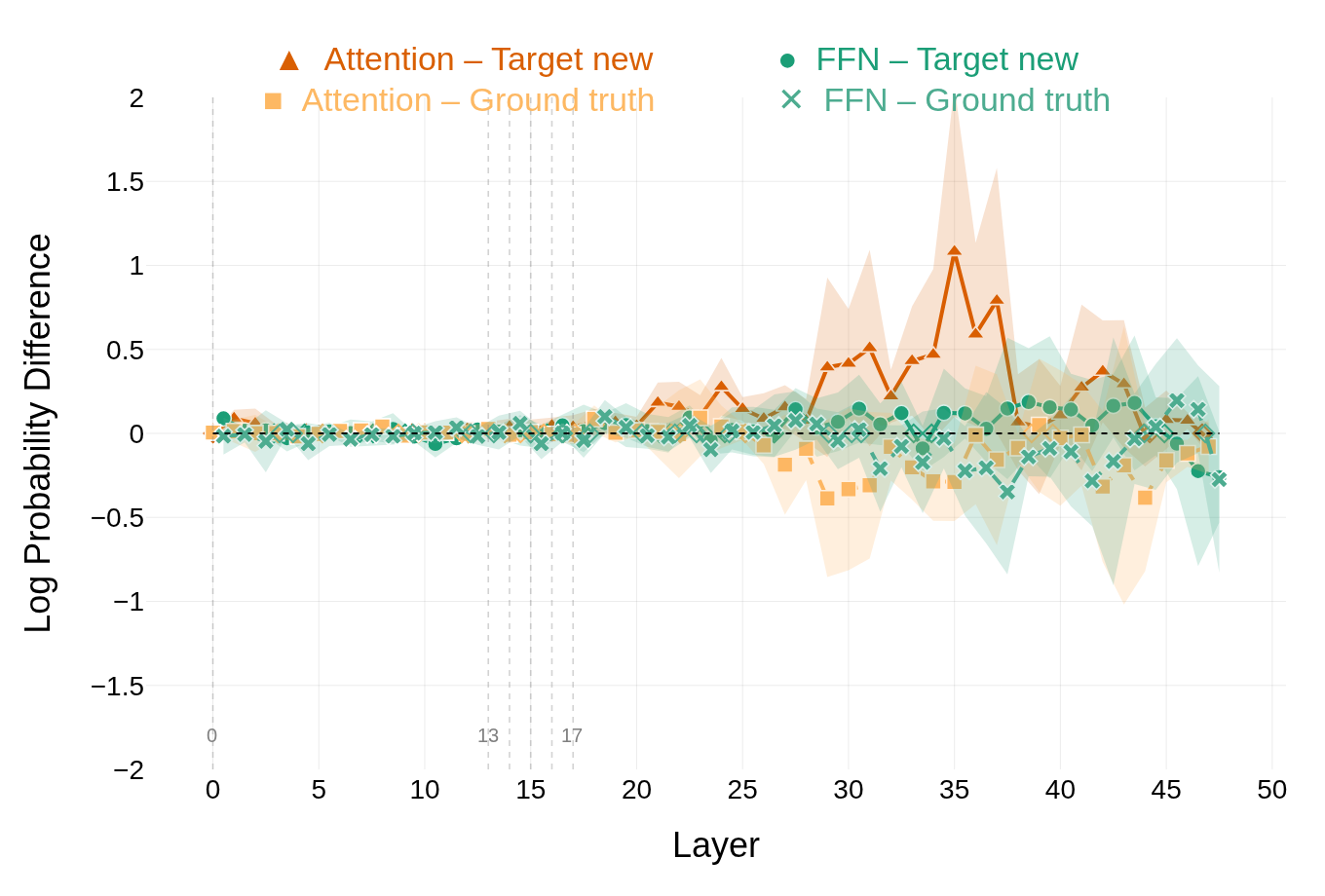} &
\includegraphics[width=8cm]{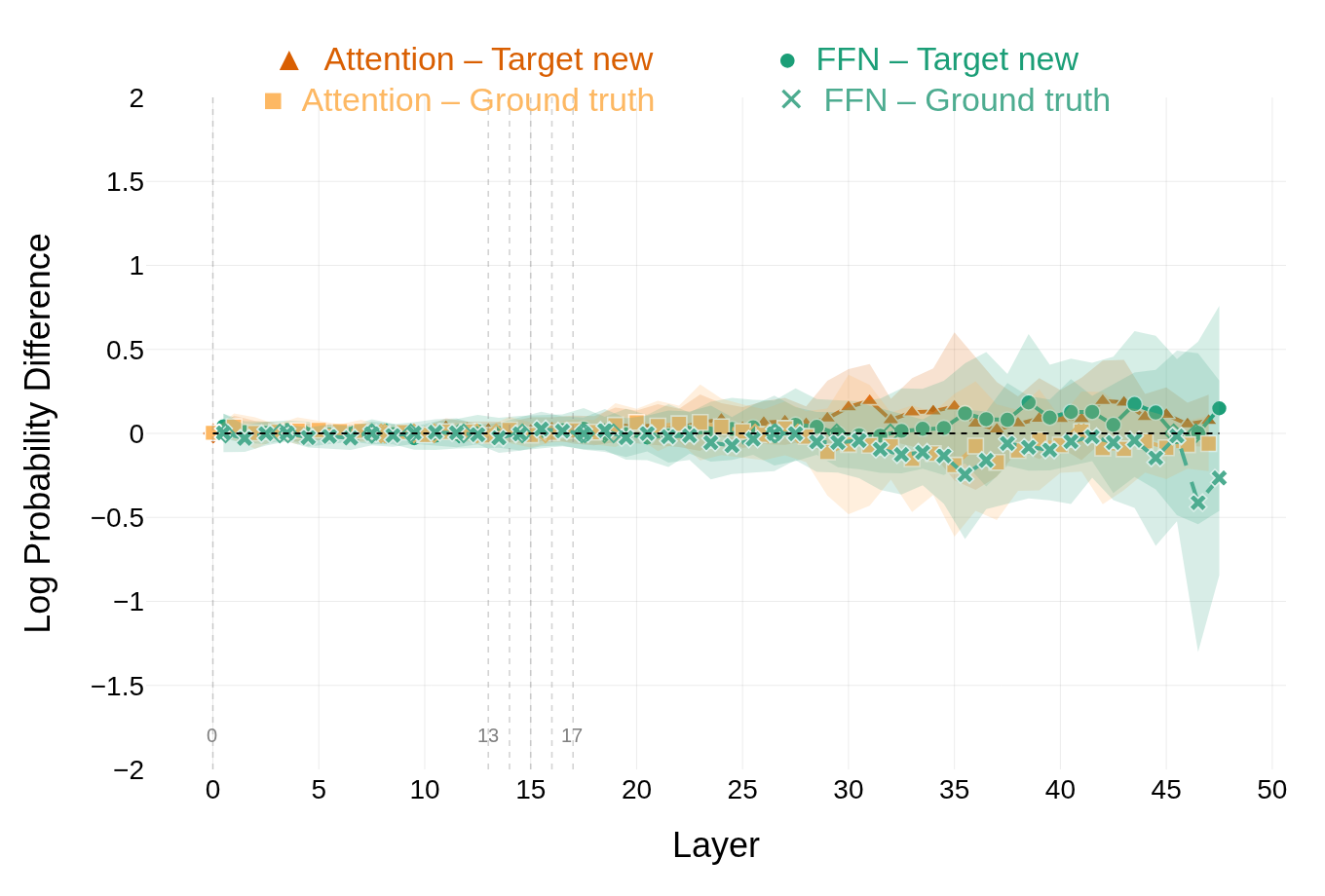} \\
(a) FT Success (L0) & (b) FT Failure (L0)
\end{tabular}
\caption{\textbf{Fine-tuning success vs.\ failure on Popular (GPT2-XL).}}
\label{fig:pop-gpt2-ft}
\end{figure*}

Figures~\ref{fig:pop-gpt2-rome}, \ref{fig:pop-gpt2-ike}, \ref{fig:pop-gpt2-memit}, and
\ref{fig:pop-gpt2-ft} show attribution patterns for successful and failed
edits on the Popular dataset for GPT2-XL.

\subsubsection{LLaMA2-7B}

\begin{figure*}[t]
\centering
\begin{tabular}{cc}
\includegraphics[width=8cm]{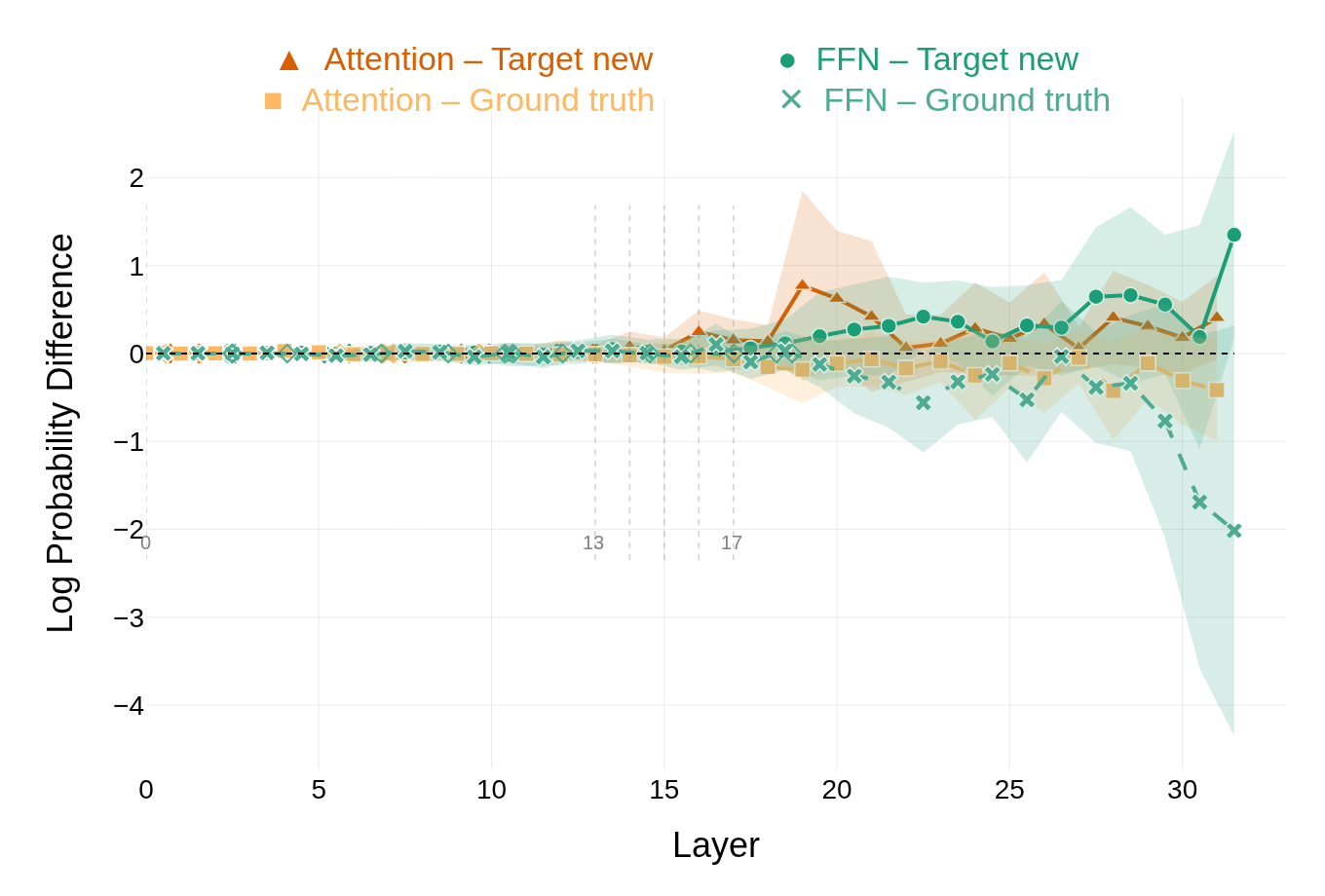} &
\includegraphics[width=8cm]{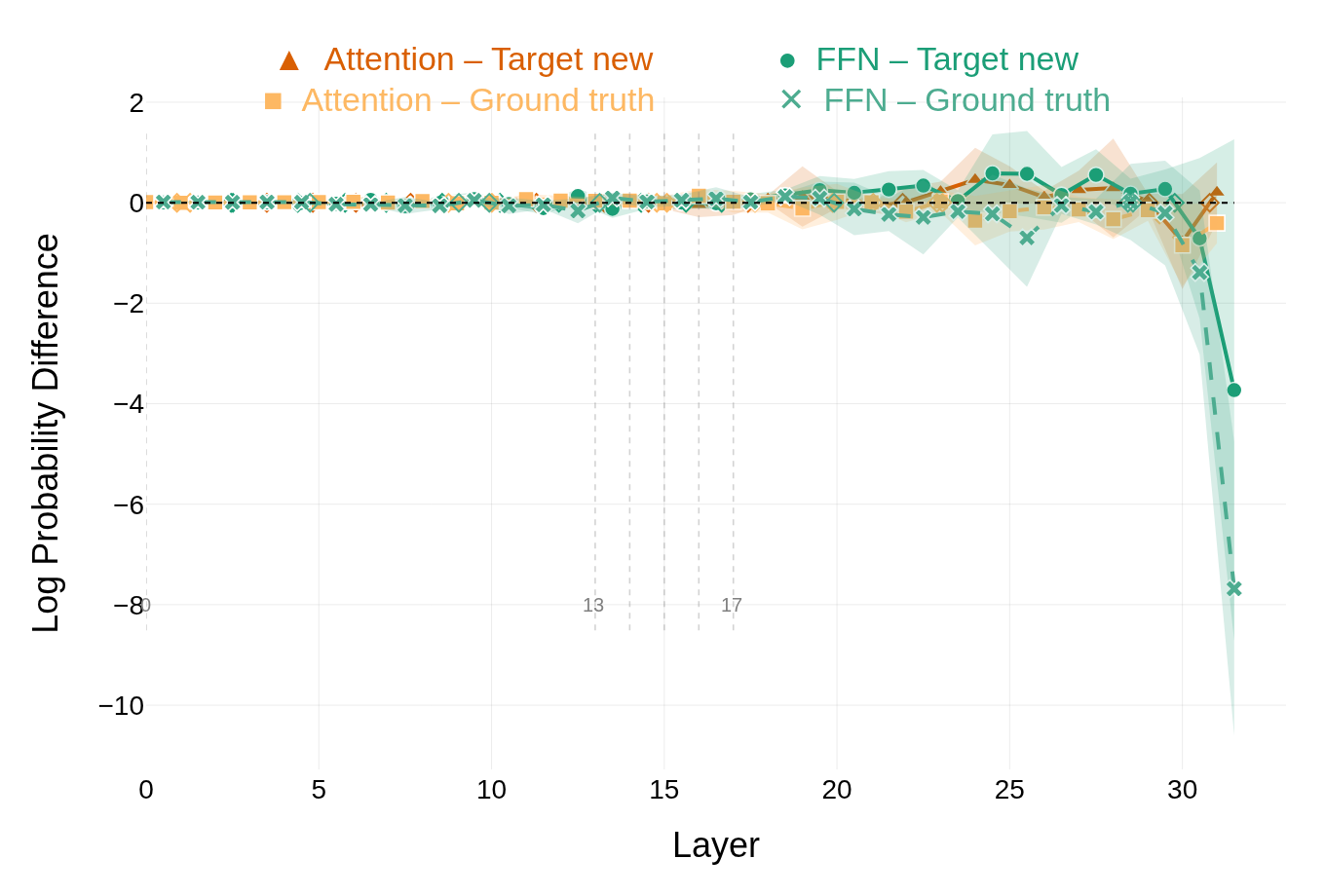} \\
(a) Success (L5) & (b) Failure (L5)
\end{tabular}
\caption{\textbf{ROME success vs.\ failure on Popular (LLaMA2-7B).}}
\label{fig:pop-llama-rome}
\end{figure*}

\begin{figure*}[t]
\centering
\begin{tabular}{cc}
\includegraphics[width=8cm]{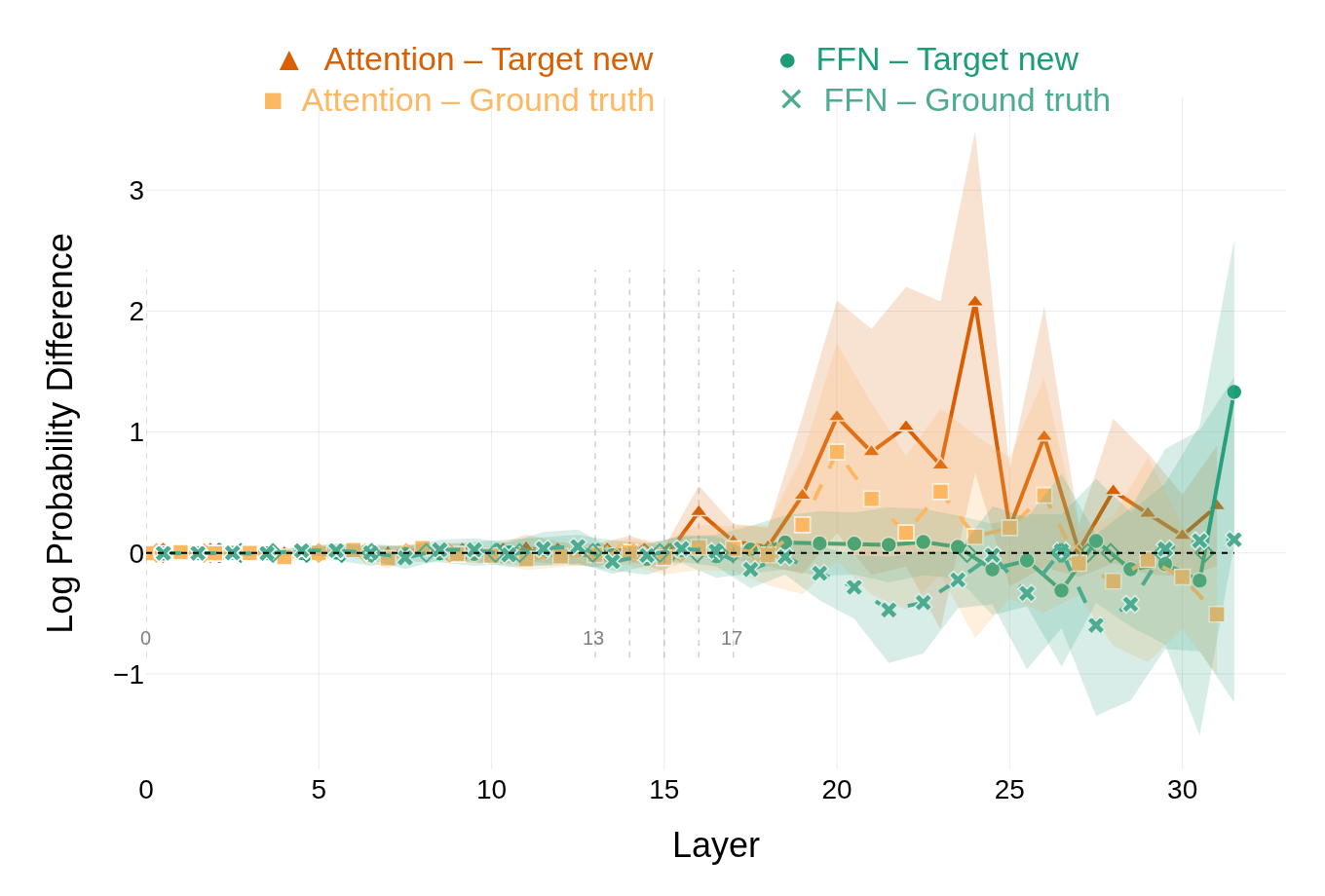} &
\includegraphics[width=8cm]{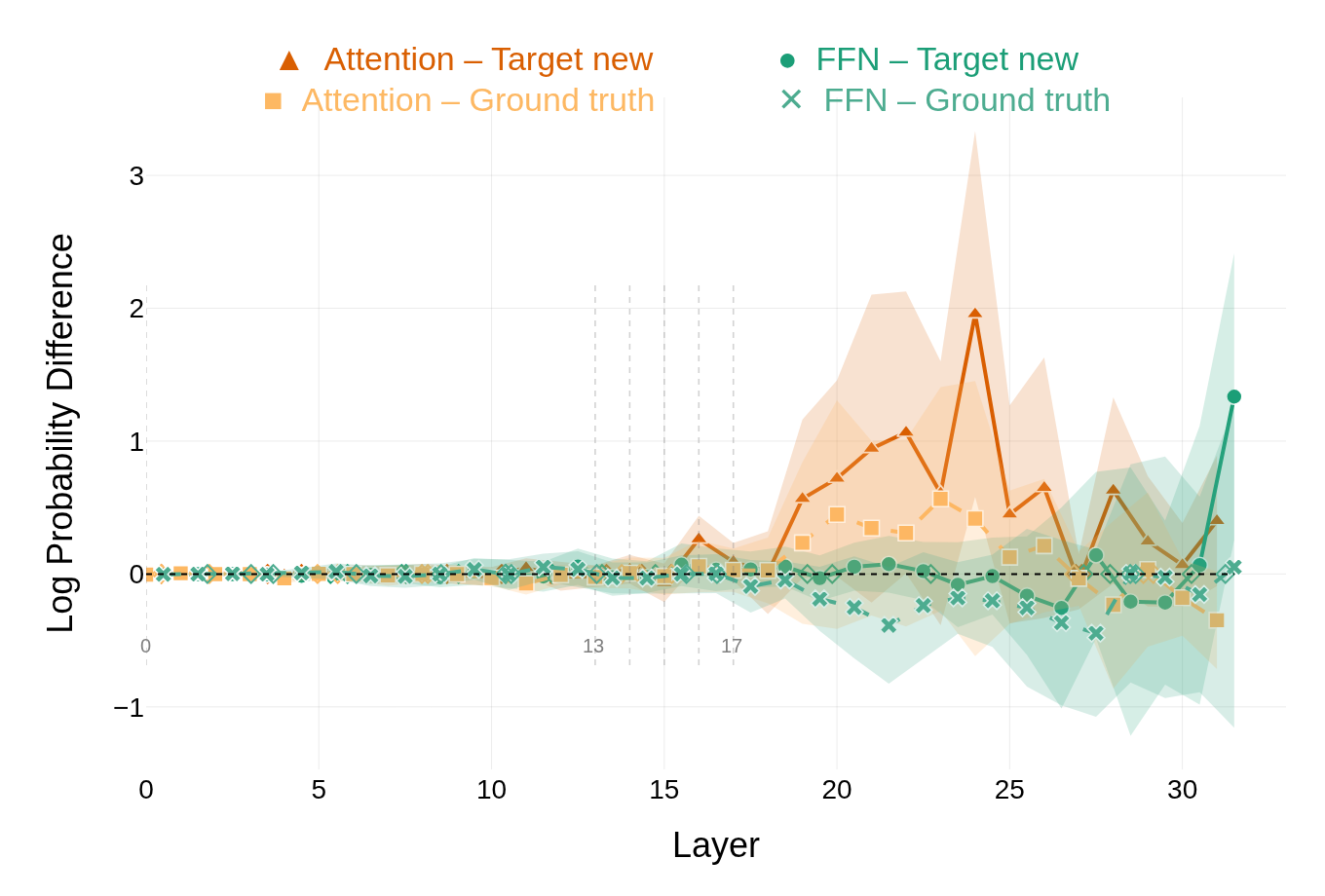} \\
(a) Success (ORI) & (b) Failure (ORI)
\end{tabular}
\caption{\textbf{IKE success vs.\ failure on Popular (LLaMA2-7B).}}
\label{fig:pop-llama-ike}
\end{figure*}

\begin{figure*}[t]
\centering
\begin{tabular}{cc}
\includegraphics[width=8cm]{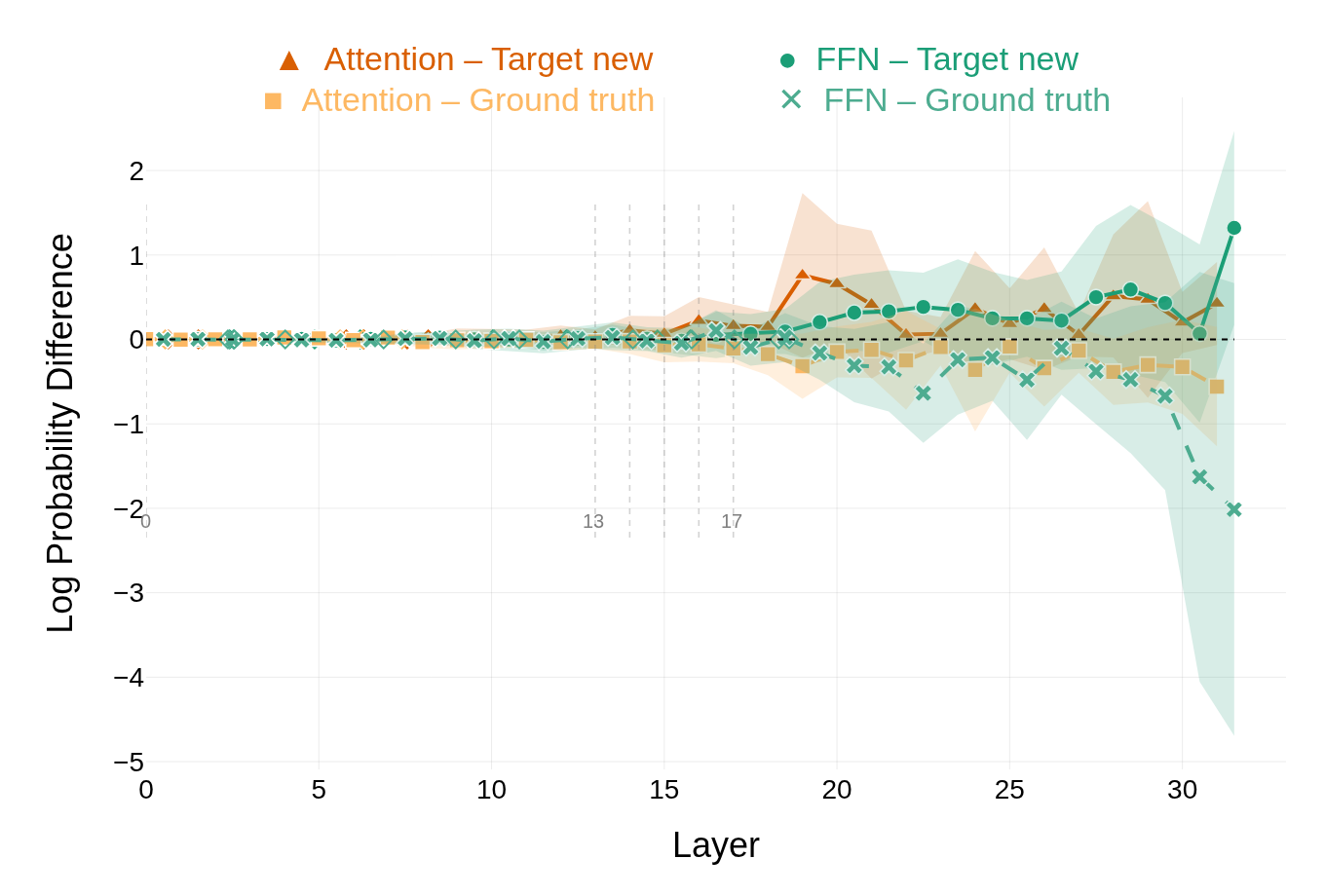} &
\includegraphics[width=8cm]{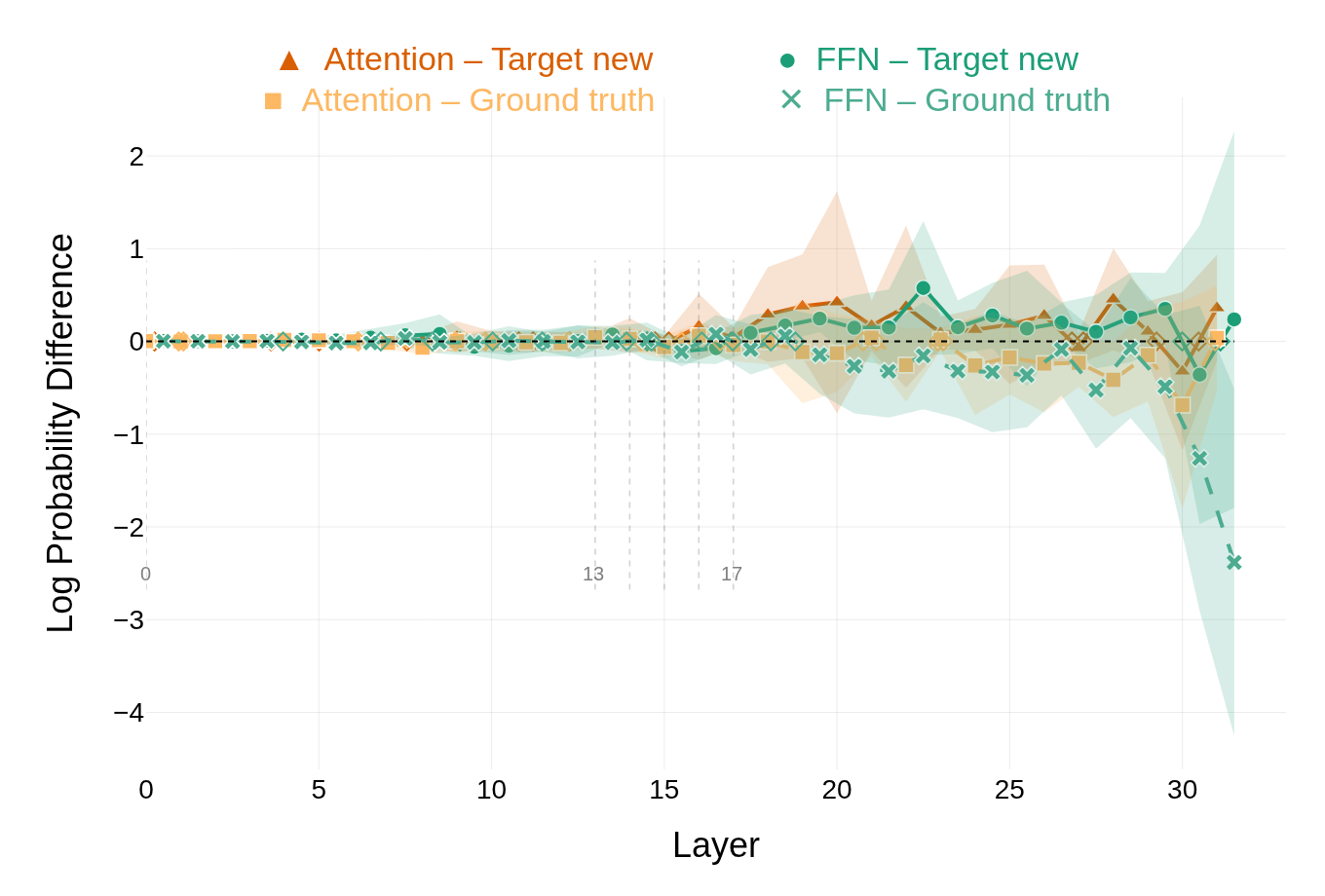} \\
(a) Success (L4--8) & (b) Failure (L4--8)
\end{tabular}
\caption{\textbf{MEMIT success vs.\ failure on Popular (LLaMA2-7B).}}
\label{fig:pop-llama-memit}
\end{figure*}

\begin{figure*}[t]
\centering
\begin{tabular}{cc}
\includegraphics[width=8cm]{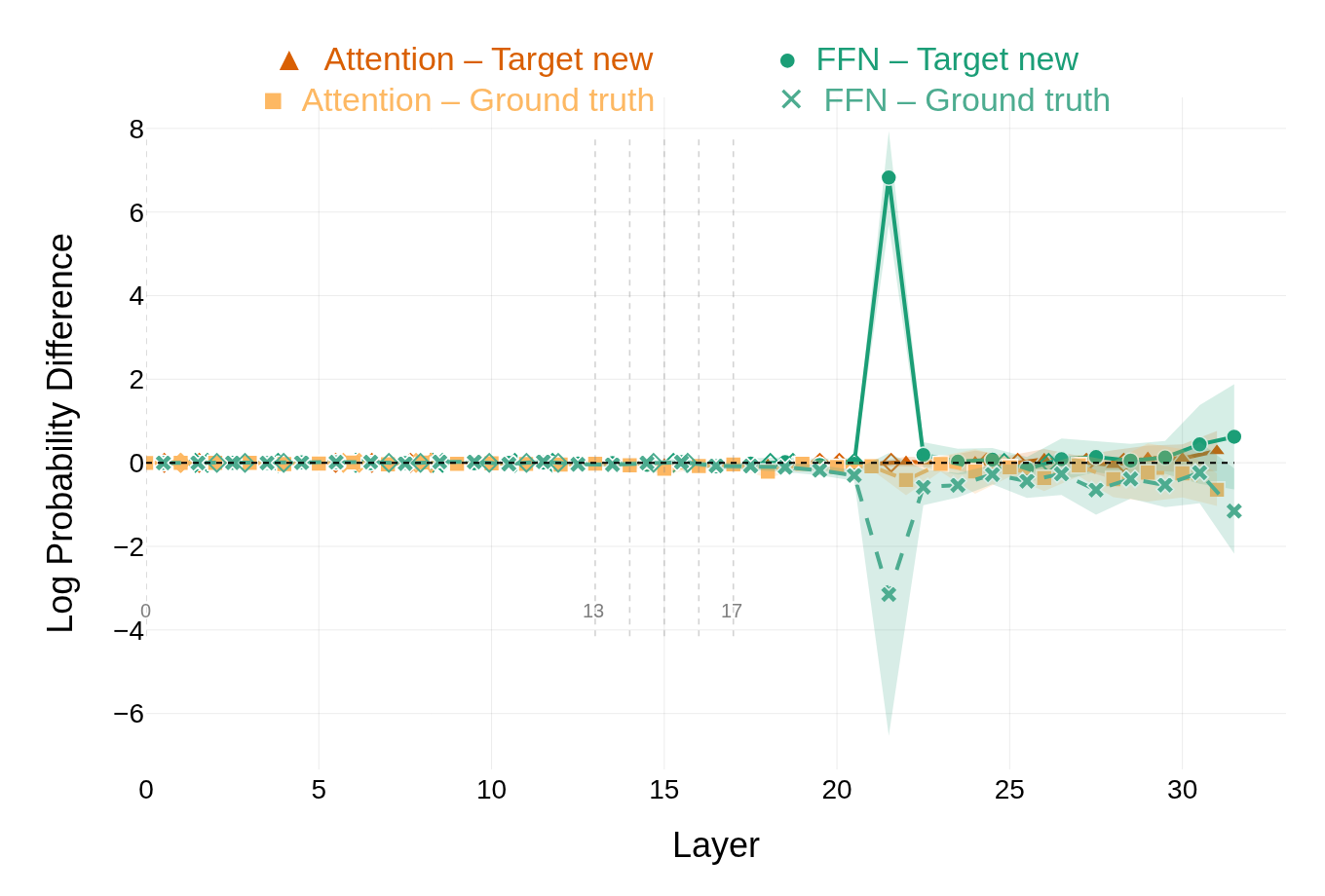} &
\includegraphics[width=8cm]{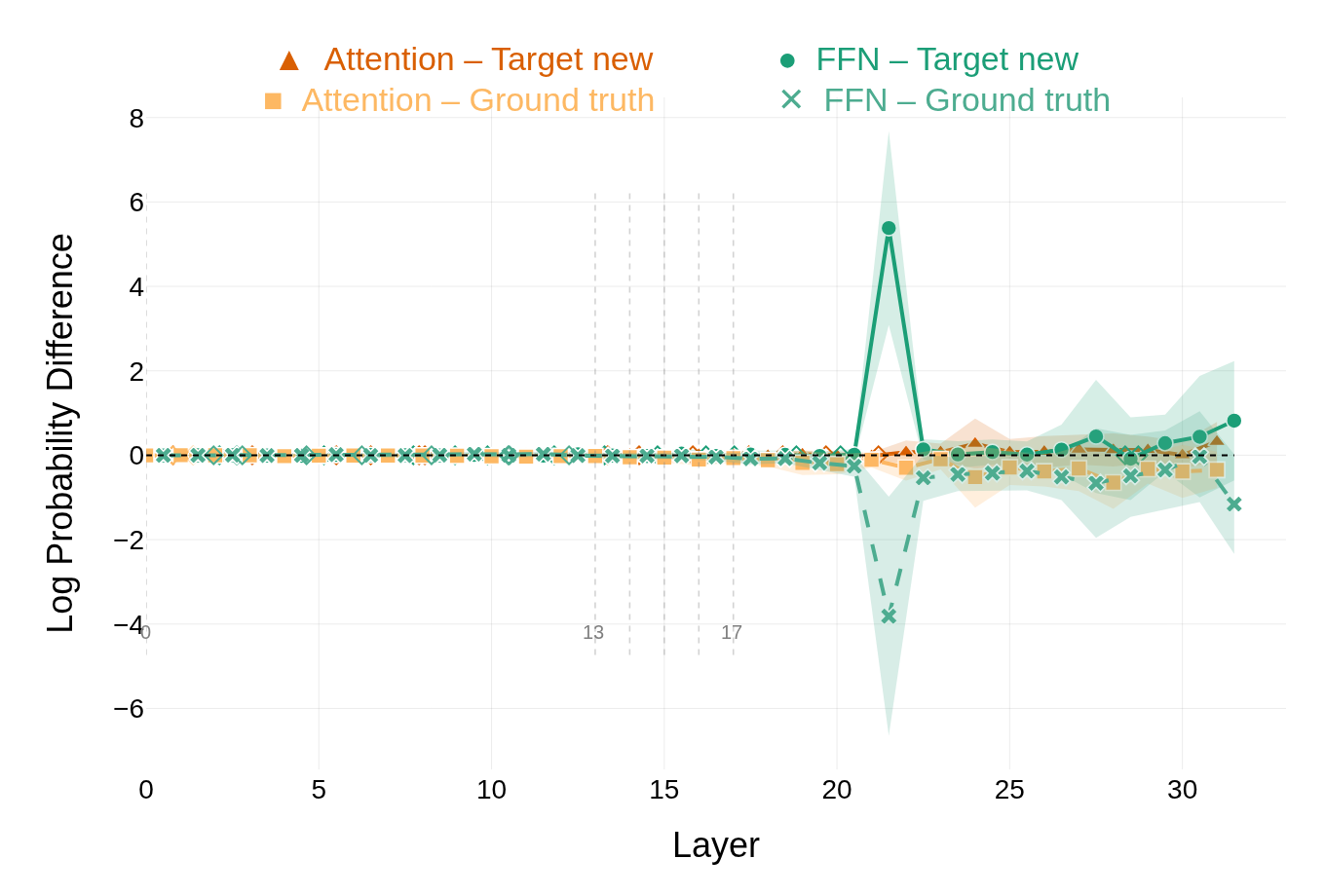} \\
(a) FT Success (L21) & (b) FT Failure (L21)
\end{tabular}
\caption{\textbf{Fine-tuning success vs.\ failure on Popular (LLaMA2-7B).}}
\label{fig:pop-llama-ft}
\end{figure*}

Figures~\ref{fig:pop-llama-rome}, \ref{fig:pop-llama-ike},
\ref{fig:pop-llama-memit}, and \ref{fig:pop-llama-ft}
show attribution patterns for successful and failed edits on the Popular
dataset for LLaMA2-7B.

\section{Additional Steering Results}
\label{app:full-results}

\subsection{CounterFact}

\subsubsection{GPT2-XL}
Tables~\ref{tab:attn_windows_cf_small}--\ref{tab:cf_single_layers_20_30} report additional CounterFact results on GPT2-XL, including attention-window ablations, multi-layer component comparisons, and selected early-to-mid single-layer sweeps. Overall, the results support attribution-aligned mid-to-late attention-residual and residual interventions, while early-layer and FFN-only variants remain weak.

\begin{table}[t]
\centering
\caption{Attention steering on CounterFact for GPT2-XL: moving a fixed-size window. Attribution-guided mid layers (15--40, 20--36) clearly outperform early or very-late windows.}
\label{tab:attn_windows_cf_small}
\small
\setlength{\tabcolsep}{3pt}
\begin{tabularx}{\columnwidth}{@{}Y l c c c c c c@{}}
\toprule
\textbf{Setting} & \textbf{Range} & \textbf{Acc} & \textbf{Gen} & \textbf{Spec} & \textbf{DI} & \textbf{DII} & \textbf{Avg}\\
\midrule
26-layer & 0--25  & 0.16 & 0.08 & 0.99 & 1.00 & 0.47 & 0.54 \\
\rowcolor{gray!12}
26-layer & 15--40 & 0.98 & 0.84 & 0.87 & 1.00 & 0.98 & 0.93 \\
26-layer & 22--47 & 0.88 & 0.71 & 0.87 & 0.99 & 0.96 & 0.88 \\
\midrule
17-layer & 0--16  & 0.01 & 0.02 & 1.00 & 0.22 & 0.02 & 0.25 \\
\rowcolor{gray!12}
17-layer & 20--36 & 0.86 & 0.73 & 0.88 & 0.98 & 0.93 & 0.88 \\
17-layer & 31--47 & 0.21 & 0.25 & 0.94 & 0.82 & 0.78 & 0.60 \\
\bottomrule
\end{tabularx}
\end{table}

\begin{table}[t]
\centering
\caption{Multi-layer steering on CounterFact (GPT2-XL), 26-layer windows.}
\label{tab:cf_26_windows_allpca}
\small
\setlength{\tabcolsep}{3pt}
\begin{tabularx}{\columnwidth}{@{}l c c c c c c c@{}}
\toprule
\textbf{Comp} & \textbf{PCA} 
& \textbf{Acc} & \textbf{Gen} & \textbf{Spec} & \textbf{DI} & \textbf{DII} & \textbf{Avg}\\
\midrule
\multicolumn{8}{c}{\textit{Range 0--25}}\\
\cmidrule(lr){1-8}
ATTN      & 128  & 0.158 & 0.079 & 0.992 & 1.000 & 0.467 & 0.539\\
ATTN      & 512  & 0.183 & 0.075 & 0.987 & 0.925 & 0.533 & 0.541\\
ATTN-res  & 128  & 0.092 & 0.075 & 0.915 & 0.708 & 0.217 & 0.401\\
ATTN-res  & 512  & 0.217 & 0.233 & 0.897 & 0.858 & 0.783 & 0.598\\
FFN       & 128  & 0.008 & 0.004 & 0.998 & 0.008 & 0.017 & 0.207\\
FFN       & 512  & 0.008 & 0.008 & 0.999 & 0.067 & 0.008 & 0.218\\
Res       & 128  & 0.158 & 0.100 & 0.994 & 1.000 & 0.408 & 0.532\\
Res       & 512  & 0.183 & 0.079 & 0.992 & 0.975 & 0.483 & 0.543\\
\midrule
\multicolumn{8}{c}{\textit{Range 15--40}}\\
\cmidrule(lr){1-8}
ATTN      & 128  & 0.983 & 0.838 & 0.871 & 1.000 & 0.975 & 0.933\\
ATTN      & 512  & 0.983 & 0.792 & 0.872 & 0.942 & 0.900 & 0.898\\
ATTN-res  & 128  & 0.983 & 0.917 & 0.871 & 0.975 & 0.967 & 0.943\\
ATTN-res  & 512  & 0.992 & 0.925 & 0.871 & 1.000 & 1.000 & 0.958\\
FFN       & 128  & 0.033 & 0.029 & 0.992 & 0.158 & 0.208 & 0.284\\
FFN       & 512  & 0.033 & 0.025 & 0.994 & 0.217 & 0.175 & 0.289\\
Res       & 128  & 0.983 & 0.892 & 0.873 & 1.000 & 1.000 & 0.950\\
Res       & 512  & 0.983 & 0.892 & 0.873 & 0.983 & 1.000 & 0.946\\
\midrule
\multicolumn{8}{c}{\textit{Range 22--47}}\\
\cmidrule(lr){1-8}
ATTN      & 128  & 0.883 & 0.713 & 0.872 & 0.992 & 0.958 & 0.884\\
ATTN      & 512  & 0.892 & 0.704 & 0.883 & 0.925 & 0.908 & 0.862\\
ATTN-res  & 128  & 0.992 & 0.925 & 0.871 & 0.975 & 0.983 & 0.949\\
ATTN-res  & 512  & 0.992 & 0.925 & 0.871 & 1.000 & 0.992 & 0.956\\
FFN       & 128  & 0.033 & 0.029 & 0.995 & 0.183 & 0.067 & 0.262\\
FFN       & 512  & 0.042 & 0.025 & 0.994 & 0.183 & 0.100 & 0.269\\
Res       & 128  & 0.875 & 0.763 & 0.893 & 0.975 & 0.917 & 0.884\\
Res       & 512  & 0.967 & 0.817 & 0.880 & 1.000 & 0.992 & 0.931\\
\bottomrule
\end{tabularx}
\end{table}

\begin{table}[t]
\centering
\caption{Multi-layer steering on CounterFact (GPT2-XL), 17-layer windows.}
\label{tab:cf_17_windows_allpca}
\small
\setlength{\tabcolsep}{3pt}
\begin{tabularx}{\columnwidth}{@{}l c c c c c c c@{}}
\toprule
\textbf{Comp} & \textbf{PCA}
& \textbf{Acc} & \textbf{Gen} & \textbf{Spec} & \textbf{DI} & \textbf{DII} & \textbf{Avg}\\
\midrule
\multicolumn{8}{c}{\textit{Range 0--16}}\\
\cmidrule(lr){1-8}
ATTN      & 128  & 0.008 & 0.017 & 1.000 & 0.217 & 0.017 & 0.252\\
ATTN      & 512  & 0.008 & 0.013 & 1.000 & 0.167 & 0.008 & 0.239\\
ATTN-res  & 128  & 0.025 & 0.008 & 0.969 & 0.533 & 0.017 & 0.310\\
ATTN-res  & 512  & 0.017 & 0.017 & 0.978 & 0.283 & 0.092 & 0.277\\
FFN       & 128  & 0.008 & 0.013 & 1.000 & 0.000 & 0.033 & 0.211\\
FFN       & 512  & 0.008 & 0.008 & 1.000 & 0.033 & 0.008 & 0.212\\
Res       & 128  & 0.008 & 0.013 & 1.000 & 0.308 & 0.017 & 0.269\\
Res       & 512  & 0.008 & 0.017 & 1.000 & 0.142 & 0.017 & 0.237\\
\midrule
\multicolumn{8}{c}{\textit{Range 20--36}}\\
\cmidrule(lr){1-8}
ATTN      & 128  & 0.858 & 0.733 & 0.875 & 0.983 & 0.925 & 0.875\\
ATTN      & 512  & 0.858 & 0.717 & 0.877 & 0.892 & 0.850 & 0.839\\
ATTN-res  & 128  & 0.983 & 0.896 & 0.871 & 0.950 & 0.958 & 0.932\\
ATTN-res  & 512  & 0.983 & 0.921 & 0.871 & 1.000 & 0.983 & 0.952\\
FFN       & 128  & 0.025 & 0.033 & 0.995 & 0.200 & 0.183 & 0.287\\
FFN       & 512  & 0.025 & 0.025 & 0.996 & 0.233 & 0.133 & 0.283\\
Res       & 128  & 0.933 & 0.875 & 0.887 & 1.000 & 0.992 & 0.937\\
Res       & 512  & 0.958 & 0.888 & 0.874 & 0.983 & 1.000 & 0.941\\
\midrule
\multicolumn{8}{c}{\textit{Range 31--47}}\\
\cmidrule(lr){1-8}
ATTN      & 128  & 0.208 & 0.254 & 0.935 & 0.817 & 0.775 & 0.598\\
ATTN      & 512  & 0.225 & 0.267 & 0.933 & 0.783 & 0.758 & 0.593\\
ATTN-res  & 128  & 1.000 & 0.925 & 0.871 & 0.983 & 0.992 & 0.954\\
ATTN-res  & 512  & 1.000 & 0.925 & 0.871 & 1.000 & 0.992 & 0.958\\
FFN       & 128  & 0.025 & 0.025 & 0.996 & 0.058 & 0.033 & 0.228\\
FFN       & 512  & 0.017 & 0.021 & 0.995 & 0.067 & 0.033 & 0.227\\
Res       & 128  & 0.883 & 0.758 & 0.894 & 0.983 & 0.908 & 0.885\\
Res       & 512  & 0.967 & 0.817 & 0.880 & 1.000 & 0.992 & 0.931\\
\bottomrule
\end{tabularx}
\end{table}

\begin{table}[t]
\centering
\caption{Single-layer steering on CounterFact (GPT2-XL), layers 0--15.}
\label{tab:cf_single_layers_0_15}
\small
\setlength{\tabcolsep}{3pt}
\begin{tabularx}{\columnwidth}{@{}l c c c c c c c@{}}
\toprule
\textbf{Comp} & \textbf{PCA}
& \textbf{Acc} & \textbf{Gen} & \textbf{Spec} & \textbf{DI} & \textbf{DII} & \textbf{Avg}\\
\midrule
\multicolumn{8}{c}{\textit{Layer 0}}\\
\cmidrule(lr){1-8}
ATTN      & 1600 & 0.008 & 0.008 & 1.000 & 0.067 & 0.000 & 0.217\\
ATTN      & 128  & 0.008 & 0.008 & 1.000 & 0.067 & 0.000 & 0.217\\
ATTN      & 512  & 0.008 & 0.008 & 1.000 & 0.067 & 0.000 & 0.217\\
ATTN-res  & 1600 & 0.008 & 0.008 & 1.000 & 0.067 & 0.000 & 0.217\\
ATTN-res  & 128  & 0.008 & 0.008 & 0.998 & 0.000 & 0.000 & 0.203\\
ATTN-res  & 512  & 0.008 & 0.008 & 1.000 & 0.000 & 0.000 & 0.203\\
FFN       & 1600 & 0.008 & 0.008 & 1.000 & 0.067 & 0.000 & 0.217\\
FFN       & 128  & 0.008 & 0.008 & 1.000 & 0.000 & 0.000 & 0.203\\
FFN       & 512  & 0.008 & 0.008 & 1.000 & 0.000 & 0.000 & 0.203\\
Res       & 1600 & 0.008 & 0.008 & 1.000 & 0.067 & 0.000 & 0.217\\
Res       & 128  & 0.008 & 0.008 & 1.000 & 0.000 & 0.000 & 0.203\\
Res       & 512  & 0.008 & 0.008 & 1.000 & 0.000 & 0.000 & 0.203\\
\midrule
\multicolumn{8}{c}{\textit{Layer 5}}\\
\cmidrule(lr){1-8}
ATTN      & 1600 & 0.008 & 0.008 & 1.000 & 0.067 & 0.000 & 0.217\\
ATTN      & 128  & 0.008 & 0.008 & 1.000 & 0.067 & 0.000 & 0.217\\
ATTN      & 512  & 0.008 & 0.008 & 1.000 & 0.067 & 0.000 & 0.217\\
ATTN-res  & 1600 & 0.008 & 0.013 & 0.999 & 0.058 & 0.000 & 0.216\\
ATTN-res  & 128  & 0.008 & 0.004 & 0.998 & 0.000 & 0.000 & 0.202\\
ATTN-res  & 512  & 0.008 & 0.008 & 0.999 & 0.000 & 0.000 & 0.203\\
FFN       & 1600 & 0.008 & 0.008 & 1.000 & 0.067 & 0.000 & 0.217\\
FFN       & 128  & 0.008 & 0.008 & 1.000 & 0.075 & 0.000 & 0.218\\
FFN       & 512  & 0.008 & 0.008 & 1.000 & 0.075 & 0.000 & 0.218\\
Res       & 1600 & 0.008 & 0.013 & 0.999 & 0.067 & 0.000 & 0.217\\
Res       & 128  & 0.008 & 0.004 & 0.999 & 0.000 & 0.000 & 0.202\\
Res       & 512  & 0.008 & 0.008 & 0.999 & 0.000 & 0.000 & 0.203\\
\midrule
\multicolumn{8}{c}{\textit{Layer 10}}\\
\cmidrule(lr){1-8}
ATTN      & 1600 & 0.008 & 0.008 & 1.000 & 0.075 & 0.000 & 0.218\\
ATTN      & 128  & 0.008 & 0.008 & 1.000 & 0.067 & 0.000 & 0.217\\
ATTN      & 512  & 0.008 & 0.008 & 1.000 & 0.067 & 0.000 & 0.217\\
ATTN-res  & 1600 & 0.008 & 0.013 & 1.000 & 0.075 & 0.000 & 0.219\\
ATTN-res  & 128  & 0.008 & 0.017 & 1.000 & 0.067 & 0.017 & 0.222\\
ATTN-res  & 512  & 0.008 & 0.013 & 1.000 & 0.133 & 0.008 & 0.233\\
FFN       & 1600 & 0.008 & 0.008 & 1.000 & 0.067 & 0.000 & 0.217\\
FFN       & 128  & 0.008 & 0.008 & 1.000 & 0.075 & 0.000 & 0.218\\
FFN       & 512  & 0.008 & 0.008 & 1.000 & 0.075 & 0.000 & 0.218\\
Res       & 1600 & 0.008 & 0.013 & 1.000 & 0.067 & 0.000 & 0.218\\
Res       & 128  & 0.008 & 0.017 & 1.000 & 0.067 & 0.025 & 0.223\\
Res       & 512  & 0.008 & 0.013 & 1.000 & 0.117 & 0.017 & 0.231\\
\midrule
\multicolumn{8}{c}{\textit{Layer 15}}\\
\cmidrule(lr){1-8}
ATTN      & 1600 & 0.008 & 0.008 & 1.000 & 0.067 & 0.000 & 0.217\\
ATTN      & 128  & 0.008 & 0.008 & 1.000 & 0.075 & 0.000 & 0.218\\
ATTN      & 512  & 0.008 & 0.008 & 1.000 & 0.067 & 0.000 & 0.217\\
ATTN-res  & 1600 & 0.008 & 0.013 & 1.000 & 0.092 & 0.000 & 0.223\\
ATTN-res  & 128  & 0.008 & 0.017 & 1.000 & 0.150 & 0.017 & 0.238\\
ATTN-res  & 512  & 0.008 & 0.017 & 1.000 & 0.167 & 0.033 & 0.245\\
FFN       & 1600 & 0.008 & 0.008 & 1.000 & 0.067 & 0.000 & 0.217\\
FFN       & 128  & 0.008 & 0.008 & 1.000 & 0.067 & 0.000 & 0.217\\
FFN       & 512  & 0.008 & 0.008 & 1.000 & 0.067 & 0.000 & 0.217\\
Res       & 1600 & 0.008 & 0.013 & 1.000 & 0.083 & 0.000 & 0.221\\
Res       & 128  & 0.008 & 0.017 & 1.000 & 0.267 & 0.017 & 0.262\\
Res       & 512  & 0.008 & 0.017 & 1.000 & 0.183 & 0.033 & 0.248\\
\bottomrule
\end{tabularx}
\end{table}

\begin{table}[t]
\centering
\caption{Single-layer steering on CounterFact (GPT2-XL), layers 20--30.}
\label{tab:cf_single_layers_20_30}
\small
\setlength{\tabcolsep}{3pt}
\begin{tabularx}{\columnwidth}{@{}l c c c c c c c@{}}
\toprule
\textbf{Comp} & \textbf{PCA}
& \textbf{Acc} & \textbf{Gen} & \textbf{Spec} & \textbf{DI} & \textbf{DII} & \textbf{Avg}\\
\midrule
\multicolumn{8}{c}{\textit{Layer 20}}\\
\cmidrule(lr){1-8}
ATTN      & 1600 & 0.008 & 0.013 & 1.000 & 0.125 & 0.000 & 0.229\\
ATTN      & 128  & 0.008 & 0.013 & 1.000 & 0.142 & 0.000 & 0.233\\
ATTN      & 512  & 0.008 & 0.013 & 1.000 & 0.133 & 0.000 & 0.231\\
ATTN-res  & 1600 & 0.025 & 0.029 & 1.000 & 0.467 & 0.067 & 0.318\\
ATTN-res  & 128  & 0.025 & 0.038 & 0.999 & 0.825 & 0.042 & 0.386\\
ATTN-res  & 512  & 0.025 & 0.033 & 0.999 & 0.725 & 0.108 & 0.378\\
FFN       & 1600 & 0.008 & 0.008 & 1.000 & 0.050 & 0.000 & 0.213\\
FFN       & 128  & 0.008 & 0.008 & 1.000 & 0.025 & 0.000 & 0.208\\
FFN       & 512  & 0.008 & 0.008 & 1.000 & 0.050 & 0.000 & 0.213\\
Res       & 1600 & 0.025 & 0.029 & 1.000 & 0.450 & 0.075 & 0.316\\
Res       & 128  & 0.025 & 0.038 & 0.999 & 0.842 & 0.033 & 0.387\\
Res       & 512  & 0.025 & 0.038 & 1.000 & 0.650 & 0.117 & 0.366\\
\midrule
\multicolumn{8}{c}{\textit{Layer 25}}\\
\cmidrule(lr){1-8}
ATTN      & 1600 & 0.008 & 0.008 & 1.000 & 0.150 & 0.008 & 0.235\\
ATTN      & 128  & 0.008 & 0.008 & 1.000 & 0.192 & 0.025 & 0.247\\
ATTN      & 512  & 0.008 & 0.008 & 1.000 & 0.092 & 0.017 & 0.225\\
ATTN-res  & 1600 & 0.192 & 0.088 & 0.991 & 0.892 & 0.467 & 0.526\\
ATTN-res  & 128  & 0.167 & 0.104 & 0.994 & 1.000 & 0.425 & 0.538\\
ATTN-res  & 512  & 0.175 & 0.079 & 0.992 & 0.975 & 0.517 & 0.548\\
FFN       & 1600 & 0.008 & 0.008 & 1.000 & 0.067 & 0.000 & 0.217\\
FFN       & 128  & 0.008 & 0.008 & 1.000 & 0.067 & 0.000 & 0.217\\
FFN       & 512  & 0.008 & 0.008 & 1.000 & 0.067 & 0.000 & 0.217\\
Res       & 1600 & 0.183 & 0.088 & 0.990 & 0.825 & 0.425 & 0.502\\
Res       & 128  & 0.158 & 0.096 & 0.994 & 1.000 & 0.400 & 0.530\\
Res       & 512  & 0.167 & 0.083 & 0.992 & 0.967 & 0.492 & 0.540\\
\midrule
\multicolumn{8}{c}{\textit{Layer 30}}\\
\cmidrule(lr){1-8}
ATTN      & 1600 & 0.008 & 0.013 & 0.996 & 0.167 & 0.008 & 0.238\\
ATTN      & 128  & 0.008 & 0.008 & 0.998 & 0.125 & 0.008 & 0.230\\
ATTN      & 512  & 0.008 & 0.008 & 0.998 & 0.125 & 0.008 & 0.230\\
ATTN-res  & 1600 & 0.625 & 0.500 & 0.936 & 0.942 & 0.842 & 0.769\\
ATTN-res  & 128  & 0.517 & 0.529 & 0.956 & 1.000 & 0.925 & 0.785\\
ATTN-res  & 512  & 0.633 & 0.563 & 0.937 & 0.983 & 0.917 & 0.807\\
FFN       & 1600 & 0.008 & 0.008 & 1.000 & 0.083 & 0.008 & 0.222\\
FFN       & 128  & 0.008 & 0.008 & 1.000 & 0.100 & 0.017 & 0.227\\
FFN       & 512  & 0.008 & 0.008 & 1.000 & 0.092 & 0.017 & 0.225\\
Res       & 1600 & 0.633 & 0.488 & 0.936 & 0.950 & 0.833 & 0.768\\
Res       & 128  & 0.500 & 0.504 & 0.955 & 1.000 & 0.925 & 0.777\\
Res       & 512  & 0.608 & 0.550 & 0.935 & 0.975 & 0.892 & 0.792\\
\bottomrule
\end{tabularx}
\end{table}

\subsubsection{LLaMA2-7B}
Table~\ref{tab:cf_multi_layers_attn_res} reports additional CounterFact results on LLaMA2-7B for attribution-aligned mid-to-late window interventions.

\begin{table}[t]
\centering
\caption{Multi-layer steering on CounterFact (LLaMA2-7B), layer windows.}
\label{tab:cf_multi_layers_attn_res}
\small
\setlength{\tabcolsep}{3pt}
\begin{tabularx}{\columnwidth}{@{}l c c c c c c c@{}}
\toprule
\textbf{Comp} & \textbf{PCA}
& \textbf{Acc} & \textbf{Gen} & \textbf{Spec} & \textbf{DI} & \textbf{DII} & \textbf{Avg}\\
\midrule
\multicolumn{8}{c}{\textit{Range 13--31}}\\
\cmidrule(lr){1-8}
ATTN      & 512 & 0.942 & 0.825 & 0.888 & 0.917 & 0.950 & 0.904 \\
ATTN-res  & 512 & 0.992 & 0.913 & 0.888 & 1.000 & 0.967 & 0.952 \\
FFN       & 512 & 0.000 & 0.008 & 0.888 & 0.250 & 0.433 & 0.316 \\
Res       & 512 & 0.908 & 0.779 & 0.888 & 0.767 & 0.875 & 0.844 \\
\midrule
\multicolumn{8}{c}{\textit{Range 15--27}}\\
\cmidrule(lr){1-8}
ATTN      & 512 & 0.875 & 0.717 & 0.888 & 0.833 & 0.950 & 0.853 \\
ATTN-res  & 512 & 0.992 & 0.913 & 0.888 & 0.975 & 0.883 & 0.930 \\
FFN       & 512 & 0.008 & 0.029 & 0.888 & 0.483 & 0.608 & 0.404 \\
Res       & 512 & 0.967 & 0.904 & 0.888 & 1.000 & 1.000 & 0.952 \\
\bottomrule
\end{tabularx}
\end{table}

\subsection{Popular}

\subsubsection{GPT2-XL}
Table~\ref{tab:pop_multi_windows} reports additional Popular results on GPT2-XL for attribution-aligned mid-to-late configurations. Although residual-only variants are competitive in some Popular settings, we treat attention-residual steering as the primary instantiation of MEGA because it most directly reflects the mechanistic hypothesis suggested by attribution: target-promoting attention contributions are integrated through the residual stream.

\begin{table}[t]
\centering
\caption{Multi-layer steering on Popular (GPT2-XL), layer windows.}
\label{tab:pop_multi_windows}
\small
\setlength{\tabcolsep}{3pt}
\begin{tabularx}{\columnwidth}{@{}l c c c c c c c@{}}
\toprule
\textbf{Comp} & \textbf{PCA}
& \textbf{Acc} & \textbf{CI} & \textbf{CII} & \textbf{RS} & \textbf{SA} & \textbf{Avg}\\
\midrule
\multicolumn{8}{c}{\textit{Range 15--40}}\\
\cmidrule(lr){1-8}
ATTN     & 128 & 0.667 & 0.275 & 0.400 & 0.382 & 0.703 & 0.485 \\
ATTN     & 512 & 0.392 & 0.250 & 0.400 & 0.441 & 0.658 & 0.428 \\
ATTN-res & 128 & 0.642 & 0.225 & 0.400 & 0.324 & 0.679 & 0.454 \\
ATTN-res & 512 & 0.817 & 0.200 & 0.450 & 0.324 & 0.648 & 0.488 \\
FFN      & 128 & 0.050 & 0.410 & 0.000 & 0.660 & 1.000 & 0.424 \\
FFN      & 512 & 0.017 & 0.375 & 0.000 & 0.714 & 1.000 & 0.421 \\
Res      & 128 & 0.817 & 0.450 & 0.250 & 0.412 & 0.682 & 0.522 \\
Res      & 512 & 0.717 & 0.650 & 0.300 & 0.456 & 0.689 & 0.562 \\
\midrule
\multicolumn{8}{c}{\textit{Range 15--47}}\\
\cmidrule(lr){1-8}
ATTN     & 128 & 0.733 & 0.300 & 0.400 & 0.338 & 0.696 & 0.494 \\
ATTN     & 512 & 0.525 & 0.250 & 0.450 & 0.353 & 0.655 & 0.447 \\
ATTN-res & 128 & 0.683 & 0.200 & 0.450 & 0.324 & 0.670 & 0.465 \\
ATTN-res & 512 & 0.817 & 0.250 & 0.450 & 0.324 & 0.640 & 0.496 \\
FFN      & 128 & 0.058 & 0.235 & 0.000 & 0.588 & 1.000 & 0.376 \\
FFN      & 512 & 0.017 & 0.276 & 0.000 & 0.681 & 1.000 & 0.395 \\
Res      & 128 & 0.433 & 0.462 & 0.200 & 0.569 & 0.765 & 0.486 \\
Res      & 512 & 0.533 & 0.600 & 0.250 & 0.515 & 0.675 & 0.515 \\
\midrule
\multicolumn{8}{c}{\textit{Range 20--45}}\\
\cmidrule(lr){1-8}
ATTN     & 128 & 0.658 & 0.275 & 0.350 & 0.426 & 0.709 & 0.484 \\
ATTN     & 512 & 0.458 & 0.275 & 0.300 & 0.426 & 0.662 & 0.424 \\
ATTN-res & 128 & 0.692 & 0.225 & 0.450 & 0.324 & 0.667 & 0.471 \\
ATTN-res & 512 & 0.817 & 0.225 & 0.450 & 0.324 & 0.649 & 0.493 \\
FFN      & 128 & 0.067 & 0.278 & 0.000 & 0.589 & 0.000 & 0.187 \\
FFN      & 512 & 0.000 & 0.300 & 0.000 & 0.681 & 0.000 & 0.196 \\
Res      & 128 & 0.675 & 0.385 & 0.150 & 0.375 & 0.745 & 0.466 \\
Res      & 512 & 0.658 & 0.475 & 0.300 & 0.485 & 0.671 & 0.518 \\
\midrule
\multicolumn{8}{c}{\textit{Range 31--47}}\\
\cmidrule(lr){1-8}
ATTN     & 128 & 0.375 & 0.575 & 0.150 & 0.683 & 0.738 & 0.504 \\
ATTN     & 512 & 0.308 & 0.450 & 0.200 & 0.657 & 0.660 & 0.455 \\
ATTN-res & 128 & 0.725 & 0.225 & 0.450 & 0.324 & 0.674 & 0.480 \\
ATTN-res & 512 & 0.800 & 0.275 & 0.450 & 0.324 & 0.638 & 0.497 \\
FFN      & 128 & 0.000 & 0.417 & 0.000 & 0.698 & 0.938 & 0.410 \\
FFN      & 512 & 0.017 & 0.296 & 0.000 & 0.696 & 1.000 & 0.402 \\
Res      & 128 & 0.433 & 0.462 & 0.200 & 0.569 & 0.765 & 0.486 \\
Res      & 512 & 0.533 & 0.600 & 0.250 & 0.515 & 0.675 & 0.515 \\
\bottomrule
\end{tabularx}
\end{table}

\subsubsection{LLaMA2-7B}
Table~\ref{tab:pop_multi_windows_llama2} reports additional Popular results on LLaMA2-7B for attribution-aligned mid-to-late layer configurations. 

\begin{table}[t]
\centering
\caption{Multi-layer steering on Popular (LLaMA2-7B), layer windows.}
\label{tab:pop_multi_windows_llama2}
\small
\setlength{\tabcolsep}{3pt}
\begin{tabularx}{\columnwidth}{@{}l c c c c c c c@{}}
\toprule
\textbf{Comp} & \textbf{PCA}
& \textbf{Acc} & \textbf{CI} & \textbf{CII} & \textbf{RS} & \textbf{SA} & \textbf{Avg}\\
\midrule
\multicolumn{8}{c}{\textit{Range 13--31}}\\
\cmidrule(lr){1-8}
ATTN     & 512 & 0.517 & 0.082 & 0.284 & 0.354 & 0.637 & 0.375 \\
ATTN-res & 512 & 0.817 & 0.074 & 0.247 & 0.329 & 0.637 & 0.421 \\
Res      & 512 & 0.692 & 0.081 & 0.148 & 0.399 & 0.651 & 0.394 \\
\midrule
\multicolumn{8}{c}{\textit{Range 15--31}}\\
\cmidrule(lr){1-8}
ATTN     & 512 & 0.500 & 0.082 & 0.284 & 0.367 & 0.630 & 0.372 \\
ATTN-res & 512 & 0.808 & 0.074 & 0.247 & 0.329 & 0.637 & 0.419 \\
Res      & 512 & 0.692 & 0.081 & 0.148 & 0.399 & 0.651 & 0.394 \\
\bottomrule
\end{tabularx}
\end{table}

\section{Implementation Details}
\label{app:method-details}

\subsection{Knowledge Editing Baselines}
All baselines are implemented using EasyEdit (v1.0)~\citep{wang2024easyediteasytouseknowledgeediting} with 
default hyperparameters. We perform single independent edits (sequential\_edit=False), 
resetting model weights between edits. ROME targets layer 17 (GPT2-XL) and layer 5 
(LLaMA2-7B); MEMIT targets layers 13--17 (GPT2-XL) and layers 4--8 (LLaMA2-7B); 
FT fine-tunes the FFN at layer 0 (GPT2-XL) and layer 21 (LLaMA2-7B); IKE retrieves 
$k{=}16$ in-context demonstrations using \texttt{all-MiniLM-L6-v2} as the sentence 
encoder.

\paragraph{Success and failure classification.}
An edit is classified as \textit{successful} if $\text{acc}_\text{post} > 
\text{acc}_\text{pre}$ and $\text{acc}_\text{post} \geq 0.9$, and as \textit{failed} 
otherwise. Cases where the model already predicted the target correctly before editing 
are excluded as outliers. We retain around 50 successful and 50 failed cases per method 
for NLKA attribution analysis, restricting to cases where the unedited model correctly 
predicts the original ground-truth answer, ensuring a clean baseline for attribution.

\paragraph{NLKA Attribution.}
We apply the neuron-level knowledge attribution framework of 
\citet{yu2024neuronlevelknowledgeattributionlarge} to analyze post-edit model behavior. 
For each component $c \in \{\text{attn}, \text{FFN}\}$ at layer $\ell$, the 
attribution score $\Delta C_{\ell,c}(y)$ measures the log-probability increase of 
token $y$ when the component's output is added to the residual stream baseline at 
the last token position. We track two target tokens per case: \texttt{target\_new} 
(the intended edit target) and \texttt{ground\_truth} (the original answer), and 
average scores across 50 successful and 50 failed cases per method. For IKE, since no model parameters are modified, we instead apply attribution 
to the full in-context input: the retrieved demonstrations are prepended to the 
original prompt following the standard IKE format (i.e., 
\texttt{[ICL examples] + New Fact: [prompt + target] + Prompt: [prompt]}), 
and attribution scores are computed at the last token position of this 
extended sequence.

\subsection{MEGA Steering Method}

MEGA implements mechanism-guided activation steering based on attribution patterns identified by NLKA. For each edit, source and target activations are collected at the last token position from paraphrased prompts, projected into a 512-dimensional PCA subspace, and used to learn a linear optimal-transport map (PythonOT \texttt{LinearTransport}) between the two activation distributions. During inference, the current activation is transformed through this learned map and injected back into selected components via forward hooks, gated by a scope detector that determines whether the input prompt falls within the edit scope.

\paragraph{Multi-layer steering.}
Unlike single-layer steering approaches, MEGA supports interventions across
layer ranges identified through attribution analysis. For a chosen layer set
$L$, we hook the attention module at each layer $\ell \in L$ and modify its
output so that the resulting attention-residual state aligns with the mapped
target activation. For CounterFact, we steer layers 15--40 in GPT2-XL and
13--31 in LLaMA2-7B. For Popular, we steer layers 31--47 in GPT2-XL and
13--31 in LLaMA2-7B.

\paragraph{Component support.}
Steering can be applied to multiple internal components, including attention
outputs, FFN outputs, and attention-residual representations. This allows
interventions to target the internal pathways identified as most influential
by the attribution analysis.

\paragraph{Dimensionality reduction.}
Before learning the transport map, activations are projected into a
512-dimensional PCA subspace. Operating in this reduced space improves
numerical stability and produces more consistent steering directions
compared to using the full hidden-state space.

\paragraph{Hyperparameters.}
The Euclidean prompt-distance threshold is set to 6.5 for CounterFact and
5.0 for Popular for both GPT2-XL and LLaMA2-7B. The optimal transport
regularization coefficient is set to $\lambda = 10^{-2}$ across both models.
For Popular, we generate 100 paraphrases of the edit prompt together with
10 logical implication prompts, each augmented with four Gaussian
perturbations ($\sigma = 0.1$). A fixed random seed (42) is used to ensure
reproducibility.

\paragraph{Computational setup.} GPT2-XL experiments are run on a single NVIDIA RTX A6000 (49GB VRAM). LLaMA2-7B experiments are run on a single NVIDIA A100-SXM4-80GB (80GB VRAM).

\end{document}